%% file: main.tex
\documentclass{article} 
\usepackage[final]{colm2026_conference}

\usepackage{microtype}
\usepackage{hyperref}
\usepackage{url}
\usepackage{booktabs}
\usepackage[pdftex]{graphicx}
\usepackage{multirow}
\usepackage{amsmath}
\usepackage{amssymb}
\usepackage[table]{xcolor}
\usepackage{wrapfig}
\usepackage[most]{tcolorbox}
\usepackage{xcolor}
\usepackage{tabularx}
\usepackage{enumitem}
\usepackage{array}
\usepackage{tikz}
\usetikzlibrary{fit}

\definecolor{mplC0}{HTML}{1F77B4}
\definecolor{mplC1}{HTML}{FF7F0E}
\newcommand{\imgtok}[1]{\tokbox{fill=mplC0!80}{white}{}{#1}}
\newcommand{\captok}[1]{\tokbox{fill=mplC1!80}{white}{}{#1}}
\newcommand{\greytok}[1]{\tokbox{draw=black!35, fill=black!8}{black}{}{#1}}

\newcommand{\tokbox}[4]{%
\tikz[baseline=(tok.base)]\node[
    rounded corners=2pt,
    #1,
    text=#2,
    inner xsep=4pt,
    inner ysep=0pt,
    minimum height=2.4ex,
    text height=1.6ex,
    text depth=.25ex,
    font=\ttfamily\scriptsize
] (tok) {#4};%
}

\definecolor{imgcolor}{RGB}{36,114,200}
\definecolor{capcolor}{RGB}{180,80,40}
\definecolor{neutralcolor}{RGB}{110,110,110}

\newtcolorbox{modbox}[2][]{
  enhanced,
  breakable,
  colback=gray!3,
  colframe=black!45,
  boxrule=0.5pt,
  arc=2pt,
  left=8pt,right=8pt,top=6pt,bottom=6pt,
  fonttitle=\bfseries,
  title={#2},
  colbacktitle=black!8,
  coltitle=black,
  attach boxed title to top left={xshift=2mm,yshift*=-2mm},
  boxed title style={boxrule=0pt,arc=1pt},
  #1
}
\usepackage{comment}

\usepackage{lineno}

\definecolor{darkblue}{rgb}{0, 0, 0.5}
\hypersetup{colorlinks=true, citecolor=darkblue, linkcolor=darkblue, urlcolor=darkblue}

\title{Source-Modality Monitoring in Vision-Language Models}

\author{Etha Tianze Hua, Tian Yun, Ellie Pavlick \\
Department of Computer Science, Brown University \\
\texttt{\{tianze\_hua,tian\_yun,ellie\_pavlick\}@brown.edu}
}

\begin{document}

\ifcolmsubmission
\linenumbers
\fi

\maketitle

\begin{abstract}
We define and investigate \textit{source-modality monitoring} -- the ability of multimodal models to track and communicate the input source from which pieces of information originate. We consider source-modality monitoring as an instance of the more general \textit{binding problem}, and evaluate the extent to which models exploit syntactic versus semantic signals in order to bind words like \textit{image} in a user-provided prompt to specific components of their input and context (i.e., actual images). Across experiments spanning 11 vision-language models (VLMs) performing target-modality information retrieval tasks, we find that both syntactic and semantic signals play an important role, but that the latter tend to outweigh the former in cases when modalities are highly distinct distributionally. We discuss the implications of these findings for model robustness, and in the context of increasingly multimodal agentic systems.\footnote{All resources will be available at \url{https://github.com/ethahtz/source-modality-monitoring}.} 
\end{abstract}

\input{latex/1_introduction}

\input{latex/5_related_works}
\input{latex/2_behavioral_evaluation}
\input{latex/3_encoding_modality_information}
\input{latex/6_conclusion}

\newpage

\bibliography{colm2026_conference}
\bibliographystyle{colm2026_conference}

\input{latex/7_appendix}

\end{document}

%% file: latex/1_introduction.tex
\section{Introduction}

We are moving toward a world of multimodal AI agents that must build and update world models from many sources of information and interaction \citep{li2024optimus, yang2025magmafoundationmodelmultimodal, fung2025embodied, geminiroboticsteam2025geminirobotics15pushing}. In order for humans to interact effectively with such systems, AI models need to be able to track and communicate where (i.e., from which modality) information comes from. Borrowing terminology from cognitive psychology \citep{johnson1993source, b2ad955fa8c84d91b3fc649b38e80d86}, we refer to this ability as \textit{source-modality monitoring}.

We consider source-modality monitoring as an instance of the more general \textit{binding problem} \citep{SMOLENSKY1990159}: a model must associate particular content with an abstract role, such as \textit{image} or \textit{text}, and retrieve the content corresponding to the queried role. Recent work has shown that language models solve related binding problems through abstract, symbolic-like mechanisms, for example by associating entities with their attributes \citep{feng2024how, gurarieh2025mixing} or binding values to variables \citep{wu2025how}. Given this, we might expect modern multimodal models to invoke a similar mechanism for binding inputs to modalities, especially since their inputs are often explicitly marked with modality-specific wrapper tokens such as \texttt{<image>} and \texttt{</image>}.

However, viewing source-modality monitoring as a case of purely symbolic binding misses many of the key properties. Unlike arbitrary role labels, words such as \textit{image} and \textit{text} are grounded in content with distinct distributional and semantic structure. Some kinds of inputs just \textit{are} images, regardless of whether or not they are tagged as such. As a result, when a multimodal model is asked to report ``what is in the image,'' it need not rely only on explicit modality markers; it should also be able to infer from the nature of the input itself whether or not a piece of information should bind to a word like \textit{image}. Moreover, as AI becomes increasingly multimodal and agentic, it becomes unrealistic to assume that all the ``modalities'' to which a human might refer will be explicitly tagged. Humans might ask AI for information from ``yesterday's meeting'', or  ``the discussion about dinner plans''. In such cases, AI will need to rely on contextual and distributional information about the content being bound, instead of or in addition to information available in explicit syntactic markers. 

In this paper, we ask if and how modern vision-language models (VLMs) accomplish source-modality monitoring. Source-modality monitoring can be studied at multiple levels, including mechanisms acquired over training \citep{sourceaware2024} and mechanisms deployed in context at inference time. We focus on the latter, and investigate how it is accomplished in a variety of VLMs across model sizes and preprocessing choices. We first show that many of these models perform well on a task designed to test in-context source-modality monitoring, and that models with explicit marker tokens (e.g., \texttt{<image>}\&\texttt{</image>}) generally outperform those without them. We then ask what kind of signal supports this performance. By structurally perturbing symbolic modality markers, we find that models' binding behavior is driven primarily by distributional semantic signals rather than purely symbolic ones, although explicit markers still improve selectivity. Finally, by applying optimized transformations to marker and content representations, we show that symbolic cues can nonetheless be exploited to reverse models' modality attribution. Together, these results provide a more nuanced picture of how syntactic and semantic signals combine to support binding in complex multimodal settings.

%% file: latex/5_related_works.tex
\section{Related Work}

\paragraph{Cross-Modal Conflict in VLMs}
Prior work has used inconsistent multimodal inputs to study modality bias in VLM predictions \citep{deng-etal-blind-faith, pezeshkpour-etal-2025-mixed, zhang2025modalitiesconflictunimodalreasoning}. While these studies show that VLMs can favor one modality over another depending on the model, dataset, or relative unimodal confidence, they typically use questions that remain ambiguously answerable from either modality, or focus mainly on the image side of the conflict. Our task builds on our prior work \citep{hua2025vision}, which explicitly queries both image and caption information under inconsistent inputs. We broaden that evaluation to more VLMs across scales and model families, and study the underlying signals that enable VLMs to accomplish this task.

\paragraph{Interpretability Methods for Transformer-Based Models}
Analyses in this paper adapt three standard interpretability techniques to study how VLMs keep track of the source modality of a token. First, probing is a widely used technique to test whether a feature of interest is (linearly) decodable from a model's hidden states \citep{alain2017understanding, conneau-etal-2018-cram, tenney-etal-2019-bert}, with control tasks separating genuine encoding from probe capacity \citep{hewitt-liang-2019-designing, belinkov-2022-probing}. We use a linear probe with shuffled-label controls to establish that modality identity is linearly decodable from token representations. Next, activation patching establishes causal effects by copying activations from one forward pass into another and measuring the change in model output \citep{NEURIPS202092650b2e, meng2022locating, wang2023interpretability, zhang2024towards}. Whereas patching is typically used to localize the components responsible for a behavior, we employ a freeze-remove variant -- restoring contextualized content-token activations into marker-token-ablated runs -- to test whether modality identity is written into content-token representations by the markers, rather than remaining localized at the marker positions. Finally, representation steering adds a vector to the residual stream at inference time, where vectors are either extracted from contrastive inputs \citep{li2023inferencetime, rimsky-etal-2024-steering} or, as in our case, trained with gradient descent against an explicit objective with the model's native weights frozen \citep{subramani-etal-2022-extracting, wu2024reft}. While steering is typically a control method, we use trained vectors as a localization instrument, identifying the token positions and layer depths at which modality identity is most causally accessible.





%% file: latex/2_behavioral_evaluation.tex
\section{How Well Do VLMs Bind Words to Their Inputs?}

\subsection{Target-Modality Retrieval Task}

\begin{figure}
    \centering
    \includegraphics[width=0.7\linewidth]{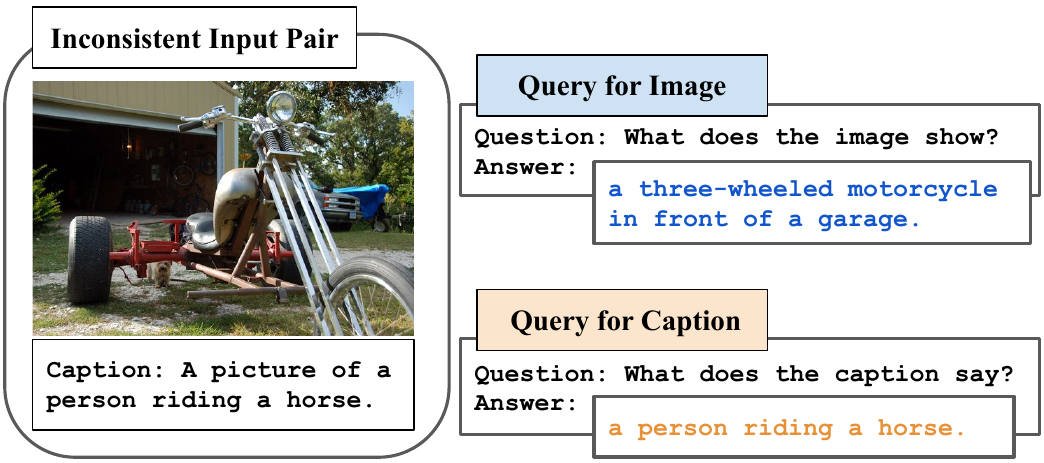}
    \caption{Example task instance.}
    \label{fig:example}
\end{figure}

We first ask simply whether VLMs are capable of associating different parts of their input (i.e., images, text) with the words that refer to those inputs (\textit{image}, \textit{caption})? We define the following task whose successful performance depends on a model's ability to monitor source modality. Each instance consists of a single in-context \textit{inconsistent} image-caption pair, and the model is prompted to retrieve information from one specified input source (e.g., ``what is in the image'' or ``what is in the caption''). We refer to the queried source as the \textit{target modality} and the other source as the \textit{non-target modality}. Figure~\ref{fig:example} shows an example, illustrating the correct answers when a model is asked to retrieve image or caption information from its input pair. We exclude the possibility that VLMs rely on input order to retrieve information by evaluating them under both input orderings -- image tokens first or text tokens first -- and report the average result across the two conditions.\footnote{We adapt this approach to all models except InstructBLIP, which has a fixed order of having image tokens at the beginning of the input sequence.}

Because the visual and textual inputs convey mutually exclusive information, answering correctly requires the model to retrieve content from the specified target modality. As the target modality is explicitly indicated, this task does not merely measure a model's overall modality bias, but instead targets at the model’s ability to associate retrieved information with its input modality thus making it suitable for assessing source-modality monitoring. 

\subsection{Datasets and Models}
To construct inconsistent image--caption pairs, we leverage existing image captioning datasets flickr30k \citep{young-etal-2014-image} and MSCOCO \citep{lin2015microsoftcococommonobjects}, and systematically sample captions that are inconsistent with the visual content of each image. We use a simple heuristic procedure to ensure that the caption paired with the image is genuinely mismatched at the semantic level. To do this, we require that the sampled caption have cosine similarity below a threshold of $0.2$ with \emph{any} of the five image's ground-truth associated captions, where similarity is computed using a BERT-style sentence embedding model (\url{sentence-transformers/all-mpnet-base-v2} \citep{mpnet})\footnote{We find the threshold of $0.2$ to be appropriate for filtering out semantically similar stimuli across modalities. For human validation of this threshold, see Appendix~\ref{app:validation_cos_threshold}.}.

We evaluate a range of 11 open-source VLMs on this task. Beyond differences in model sizes, these systems also vary along an architectural dimension that is especially relevant to our setting: whether the input preprocessing pipeline inserts explicit \textit{image wrapper tokens} that mark the boundary of image tokens. Qwen-2.5-VL, InternVL3, and Gemma-3 models \citep{Qwen2.5-VL, zhu2025internvl3, team2025gemma} are trained with these wrapper tokens, while earlier models such as LLaVA-1.5 \citep{liu2023llava}, InstructBLIP \citep{dai2023instructblip}, and LLaVA-OneVision \citep{li2025llavaonevision} do not use such markers. See Table~\ref{tab:modality_wrapping} for the image marker tokens used across model families. This variation allows us to ask whether the presence of modality-marking tokens is associated with stronger retrieval from the target modality.

\subsection{Metrics and Baselines}

\paragraph{Metrics} To measure how reliably models retrieve information from the target modality, we first identify \textit{valid predictions}, namely outputs that clearly report information from either the target modality or the non-target modality (i.e., as opposed to outputting nonsense, which models sometimes do). Then, we define \textit{selectivity} as an aggregate metric which indicates the tendency of the model to return the modality that was requested in the prompt. To obtain selectivity, we compute the difference between the frequency with which the model reports information from the target modality and the frequency with which it reports information from the non-target modality, among these valid predictions.\footnote{See Appendix~\ref{app:selectivity_formal} for the formal definitions of valid prediction rate and selectivity.} This yields a score in the range $[-1, 1]$: a score of $1$ indicates perfect retrieval from the target modality, a score of $0$ indicates no preference between the two modalities, and a score of $-1$ indicates that the model always retrieves information from the non-target modality instead.

Since we construct our input pairs from captioning datasets, model outputs are often free-form and cannot be evaluated with exact string matching. We therefore use \texttt{GPT5.4-mini} \citep{openai2026gpt54mininano} as an LLM judge to evaluate whether model outputs are grounded in the image, the caption, or neither. In Appendix~\ref{app:llm_judge_validation}, we conduct a human validation study of the LLM judge, showing that it achieves a high alignment rate with human judgment. We also include examples of model outputs and judged outcomes in Appendix~\ref{app:qual_examples_classification}. 

\paragraph{Single-Modality Baseline}
To contextualize the rate of valid responses under the inconsistent two-modality input pairs, we include a single-modality baseline in which a model receives input from only one modality and is prompted to report information from that modality. We use the same GPT model as the LLM Judge to evaluate whether the response sufficiently reports what is in its input.

\subsection{Results}
\label{sec:plain_res}

\begin{figure}[t]
    \centering
    \includegraphics[width=1\linewidth]{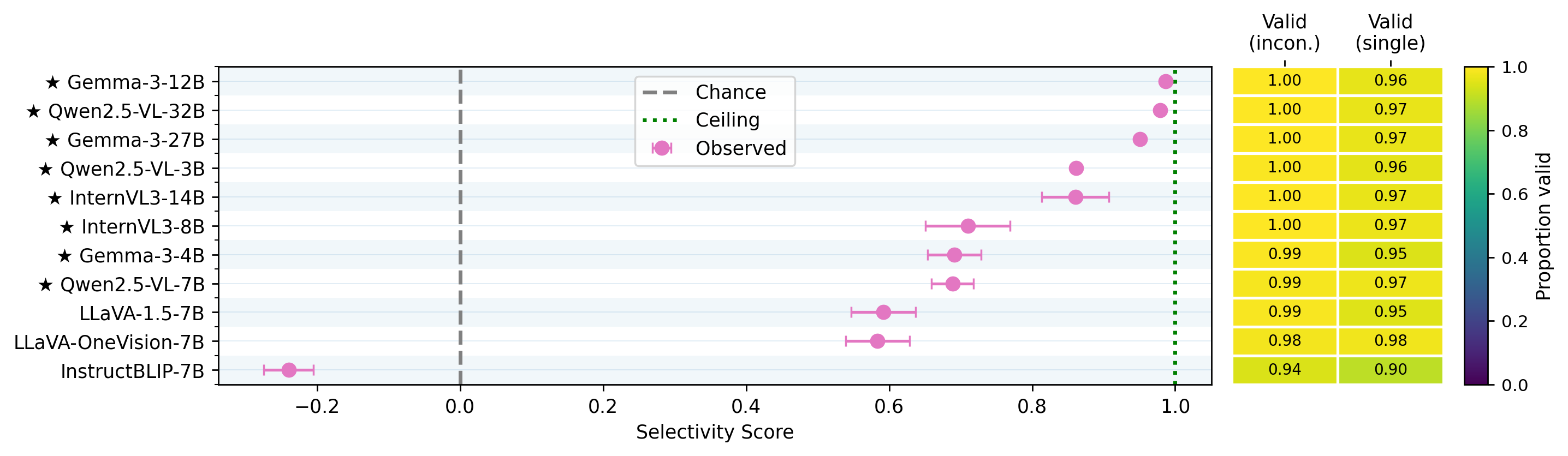}
    \caption{Left: Aggregated source-modality selectivity across VLMs and datasets. Error bars indicate standard deviation across datasets. $\bigstar$ marks VLMs with special image wrapper tokens. Right: averaged rate of valid responses under inconsistent input pairs and under single modality inputs across datasets.}
    \label{fig:overall_behavior_plots}
\end{figure}

\paragraph{Valid Responses} 
Figure~\ref{fig:overall_behavior_plots} shows the valid-response rates and selectivity scores for all 11 models in our evaluation. With the exception of InstructBLIP-7B, most models maintain a near-perfect valid-response rate under inconsistent inputs, in some cases even exceeding their single-modality baselines. One possible explanation is that inconsistent inputs encourage models to give more specific responses about the queried target modality, yielding slightly higher valid-response rates than in the single-modality setting. Still, the difference is small, suggesting that these models can report information from at least one modality with a level of specificity comparable to that seen when only a single modality is provided.

\paragraph{Modality Selectivity} 

All models, except for InstructBLIP-7B, outperform the random baseline in selectivity, indicating that they retrieve information from the target modality more often than from the non-target modality. That said, there is substantial variation between models, and many smaller models struggle to reliably return the requested information. InstructBLIP-7B shows an interesting negative selectivity, indicating it more frequently reports information from the non-target modality than the target modality. Conceptually, the worst performing model would be at chance level -- we suspect InstructBLIP is below chance level because of its fixed-order design, which puts the image tokens always before the textual tokens, that lead to a constant bias which leads to the negative selectivity. On the other end, models such as Gemma-3-12B and Qwen2.5-VL-32B perform nearly perfectly, almost achieving the ceiling performance. In general, larger models tend to perform better than smaller models; but we also observe that models with special image wrapper tokens (marked with $\bigstar$ in the figure) outperform models without such tokens. We explore this mechanism in detail in subsequent sections.

%% file: latex/3_encoding_modality_information.tex
\section{Role of Symbolic vs. Distributional Signals in Word-Input Binding}

The strong performance of VLMs which have special image wrapper tokens on the target-modality retrieval task might suggest that source-modality monitoring is driven primarily by these symbolic marker tokens. This would resemble the largely symbolic binding mechanisms previously studied in language models, such as binding entities to their attributes \citep{feng2024how, gurarieh2025mixing}.
In this section, however, we show a more nuanced picture. First, we show that VLMs are indeed capable of solving a purely symbolic binding problem when there is no distributional signal to exploit. Second, we confirm that image and text content tokens differ markedly in their representations' distributions, providing an additional/alternative cue for modality identity. Third, we show that source-modality monitoring is a hybrid syntactic-semantic mechanism: models rely in part on distributional content cues to guide retrieval, but symbolic markers help contextualize content tokens, leading to a more robust encoding of modality identity. We focus our analysis on three best-performing models from different model families from the previous section.

\newcommand{\imgseq}{%
\imgtok{IM$_1$}\,$\cdots$\,\imgtok{IM$_n$}%
}

\newcommand{\capseq}{%
\captok{a}\,\captok{person}\,\captok{riding}\,\captok{a}\,\captok{horse}%
}

\newcommand{\qseq}{%
\greytok{what}\,\greytok{is}\,\greytok{in}\,\greytok{the}\,\greytok{image/caption}\,\greytok{?}%
}

\subsection{VLMs Are Capable of Symbolic Binding}

\begin{figure}[h]
    \centering
    \includegraphics[width=1\linewidth]{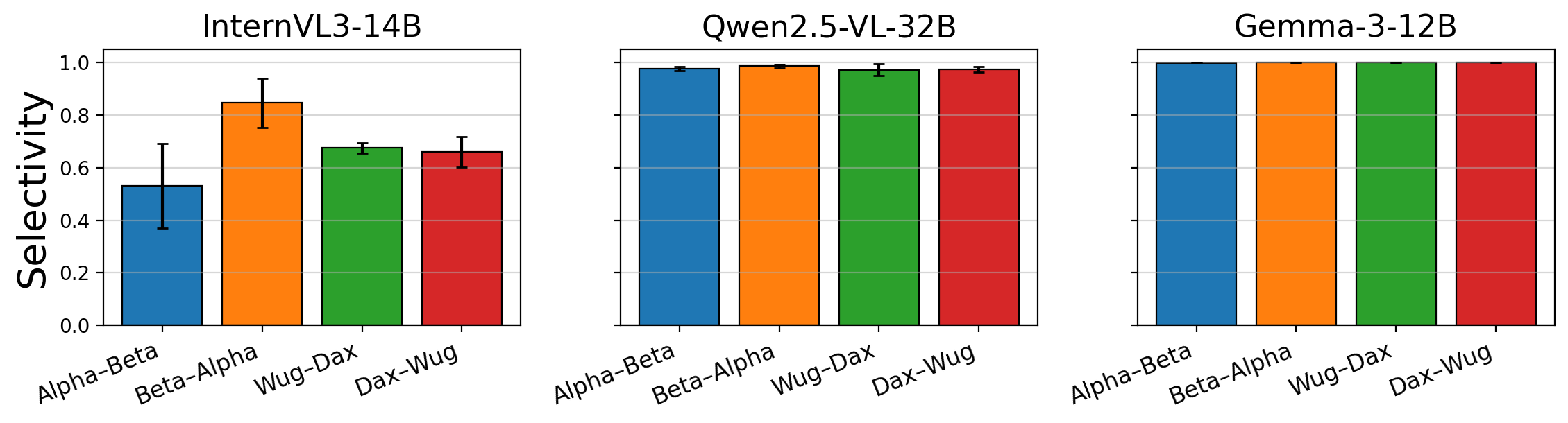}
    \caption{Performance on the purely symbolic retrieval task across four arbitrary label assignments. Each bar shows selectivity when image and caption content are associated with arbitrary labels rather than modality names. LabelA--LabelB indicates that LabelA is associated with image content and LabelB with caption content.}
    \label{fig:symbolic}
\end{figure}

We begin by asking whether models can use symbols as pure indexing devices. To isolate this capacity, we test VLMs with a simpler symbolic binding task in which the two input sources are labeled by arbitrary labels such as \textit{Alpha} and \textit{Beta} or \textit{Dax} and \textit{Wug}. Row 2 of Table~\ref{tab:marker_conditions} shows an example in which \textit{Dax} is associated with image content and \textit{Wug} with caption content. In this setting, successful performance requires the model to bind each content span to its assigned label and later retrieve the queried one, but it does not require any notion of modality. In Figure~\ref{fig:symbolic}, Qwen2.5-VL-32B and Gemma-3-12B perform nearly perfectly under this task. Although InternVL3-14B performs worse than the other two models, it still succeeds well above chance. These results show that VLMs are capable of using arbitrary symbols as indexing devices, just like their language model counterparts. 

\subsection{VLMs Encode Tokens of Different Modalities in a Linearly Separable Manner}

A second potential source of distinction comes from the distributional properties of image and text token representations. Here, we ask whether these representations are already separable before any contextual processing by the language model. Therefore, we examine them at the embedding layer of the backbone language model. To do so, we sample 200 instances from a dataset and, for each instance, sample the embedding representation of one token from each modality. We then measure cosine similarity between embedding pairs within and across modalities, and train a linear probe to distinguish image-token representations from text-token representations. To evaluate linear separability, we use stratified 3-fold cross-validation. 

Across all model-dataset pairs, we find that the averaged within-modality cosine similarity ($0.21 \pm 0.06$) is substantially higher than cross-modality cosine similarity ($0.02 \pm 0.02$). This suggests that tokens from the same modality are more similar to one another than to tokens from the other modality. Consistent with this, a linear probe can separate image and caption tokens with perfect cross-validation accuracy across all models and datasets we test, indicating that image and text representations are linearly separable to a very high degree.\footnote{See detailed results in Table~\ref{tab:distributional_separation} in the appendix.}

\begin{table}[t]
\centering
\small
\begin{tabular}{p{0.14\linewidth} p{0.80\linewidth}}
\toprule
\textbf{Condition} & \textbf{Input format} \\
\midrule
Unperturbed
&
\greytok{<image>}
\,
\imgseq
\,
\greytok{</image>}
\,
\greytok{Caption:}
\,
\capseq
\,
\greytok{.}
\\[1.2ex]

Arbitrary
&
\greytok{Dax:}
\,
\imgseq
\,
\greytok{Wug:}
\,
\capseq
\,
\greytok{.}
\\[1.2ex]

Remove
&
\imgseq
\,
\capseq
\\[1.2ex]

Swap
&
\greytok{Caption:}
\,
\imgseq
\,
\greytok{.}
\,
\greytok{<image>}
\,
\capseq
\,
\greytok{</image>}
\\
\bottomrule
\end{tabular}
\caption{Image--caption span of the prompt under the 4 marker conditions used in our experiments. In \textbf{Unperturbed}, original modality markers are preserved; in \textbf{Arbitrary}, original modality markers are replaced with arbitrary labels; in \textbf{Remove}, both image and caption markers are deleted; and in \textbf{Swap}, the image and caption markers are exchanged.}
\label{tab:marker_conditions}
\end{table}

\subsection{Structural Perturbation on Symbolic Markers}

Next, we study how symbolic and distributional signals each or together affect VLMs' ability to monitor source modalities, and retrieve from the target modality. To approach this, we systematically perturb the symbolic marker tokens by removing them or swapping them across modalities (see row 3 and 4 in Table \ref{tab:marker_conditions}). If the model uses a purely symbolic mechanism, we would expect to see a performance near chance when symbolic markers are removed, and models' predictions completely reversed when they are swapped; on the other hand, if the model relies solely on the distributional signals, we would expect these perturbations to have minimal effects on models' performance.

\begin{figure}[t]
    \centering
    \includegraphics[width=1\linewidth]{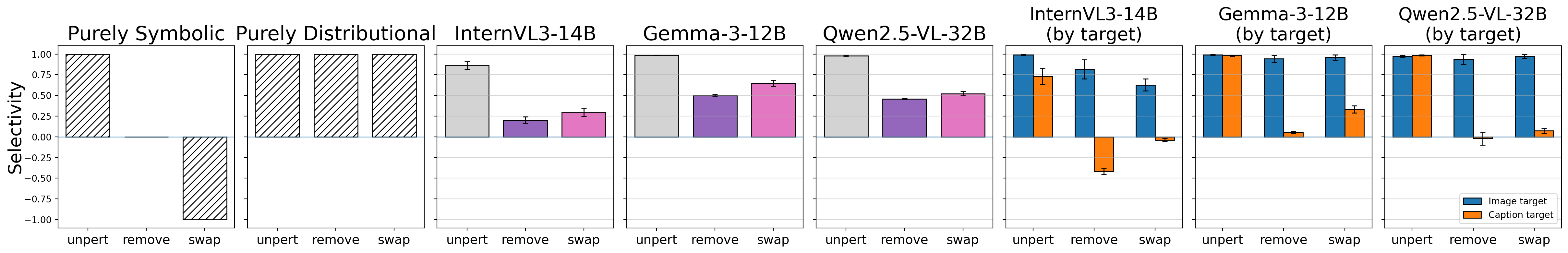}
    \caption{Selectivity under conditions where we remove or swap symbolic marker tokens, alongside purely symbolic and purely distributional expectations. Across all three VLMs, disrupting markers lowers selectivity but does not eliminate it, showing that source-modality retrieval is not reducible to symbolic cues alone. Modality-specific breakdowns show that image-target performance remains relatively unaffected, whereas caption-target performance drops sharply under perturbation, indicating that symbolic markers play a larger role in supporting caption retrieval.}
    \label{fig:perturbations}
\end{figure}

Figure~\ref{fig:perturbations} shows the result of swapping and removing symbolic marker tokens on models' performance in the target modality retrieval task. The aggregated results (subplots 3-5) shows that while removing marker tokens significantly reduces models' selectivity, it does not drop towards near-zero level, which indicates VLMs are able to use the residual signals within the distributions of image/text tokens to tell a difference between modalities. Moreover, swapping the marker tokens does not reverse models' predictions, rather it remains positive well above chance, which emphasizes that when symbolic and distributional signals conflict with each other, the model relies on the distributional signals more. 

We also include results broken down by target modality in subplots 6--8 of Figure~\ref{fig:perturbations}. We find that accuracy for reporting image content remains high across perturbation conditions, whereas accuracy for reporting caption content falls to near chance level (and even below chance for InternVL3). This sharp asymmetry suggests that VLMs learn a particularly robust way of associating the concept of ``image'' with image tokens, while relying more heavily on symbolic cues when retrieving caption information. 

One plausible reason for the asymmetry noted above is that caption tokens are embedded among other textual spans in the input, so there is nothing about them that makes them inherently ``caption'' in the way image tokens are inherently ``image.'' 

We follow up on this intuition by varying the word used to refer to the caption, and observing the effect on the above trends. In this experiment, the text source is labeled as either \textit{caption}, \textit{text}, or \textit{document}, and the query correspondingly asks what is in the image, caption, text, or document.

\begin{figure}[t]
    \centering
    \includegraphics[width=1\linewidth]{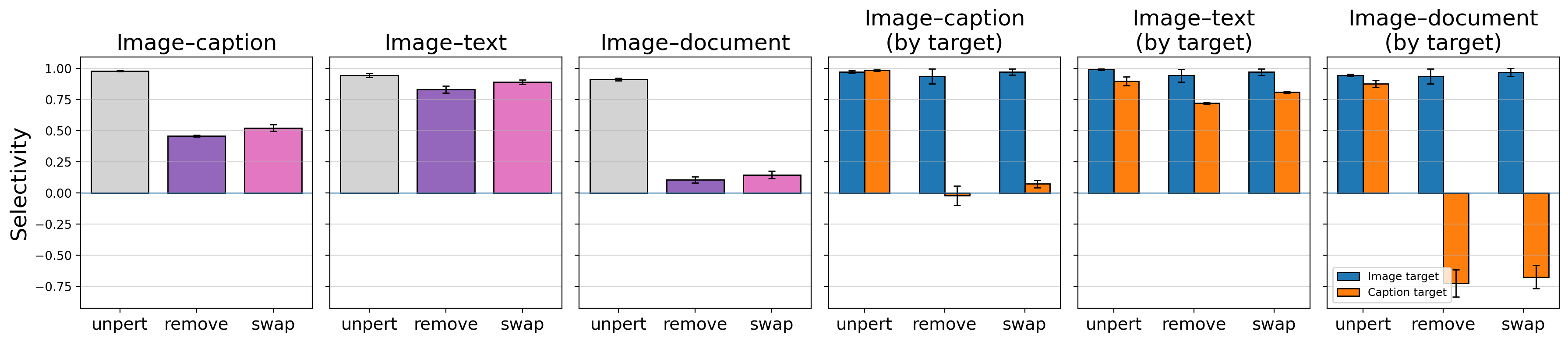}
    \caption{Selectivity across image--caption, image--text, and image--document settings under unperturbed, marker-removal, and marker-swap conditions in Qwen2.5-VL-32B. Marker perturbations have little effect in image--text, but substantially reduce performance in image--caption and image--document, where models rely heavily on markers.}
    \label{fig:different_labels}
\end{figure}

The results are shown in Figure~\ref{fig:different_labels}. Models achieve high selectivity in the unperturbed condition, showing that they can reliably retrieve information from the queried source when the source markers are present. The key difference emerges under marker perturbations, which reveal that the model relies heavily on symbolic marker tokens when the text is referred to as the \textit{document}, and relies on them very little when it is referred to as \textit{text}. This suggests an interesting and intuitive mechanism, in which models encode a notion of the type of input that is typically referred to by specific words. That is, the model is happy to bind the word \textit{text} to text in the prompt, whether or not that text is prefixed with a marker such as ``Text:''. In contrast, the model is resistant to considering an arbitrary short piece of text to be a \textit{document} unless it is explicitly labeled as such.\footnote{One plausible reason is that the category \textit{document} is intrinsically less distributionally distinct from images than \textit{text}: documents can contain heterogeneous content, including embedded images, and therefore may provide a weaker standalone signal of source identity.}

\subsection{Propagation of Symbolic Signals onto Content Tokens}

In this subsection, we investigate how symbolic and distributional signals interact, focusing on whether symbolic marker tokens help write modality identity information into content-token representations. Our hypothesis is that this happens through contextualization across layers: marker tokens influence the hidden states of downstream content tokens, causing modality identity to become embedded in those representations rather than remaining localized at the marker positions.

To isolate this effect, we first run the model on the normal input with intact markers and collect the hidden activations at the content-token positions after they have been contextualized in the presence of those markers. We then perform a second run in which the marker tokens are removed, but the contextualized content-token activations from the clean run are patched in and frozen (Figure~\ref{fig:freeze-remove-exp}). We refer to this intervention as the freeze-remove condition (Frz-RM).

\begin{figure}[t]
    \centering
    \includegraphics[width=1\linewidth]{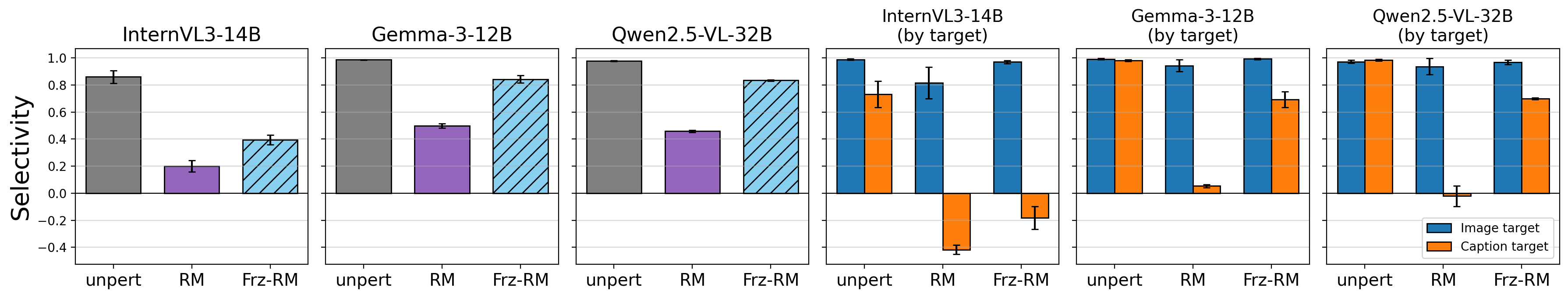}
    \caption{Selectivity under the freeze-remove intervention. Restoring contextualized content-token activations partially recovers performance relative to removing markers from the beginning, indicating that symbolic marker information is propagated into downstream content-token representations during contextualization.}
    \label{fig:freeze-remove}
\end{figure}

Figure~\ref{fig:freeze-remove} shows that symbolic information does not remain confined to the marker positions. In the Frz-RM condition, selectivity is partially restored relative to conditions in which markers are removed from the outset. This suggests that, during contextualization, information carried by the marker tokens is propagated into downstream content-token representations. Once contextualized in this way, content tokens encode not only their native distributional signals, but also the signal of modality identity inherited from the symbolic markers.

The modality-specific breakdown further shows that both image and caption content benefit from this contextualization, though the gain is much larger for caption content. For Gemma-3-12B and Qwen2.5-VL-32B, caption-target selectivity rises from near chance to around 0.6 under Frz-RM. For InternVL3-14B, the effect is weaker, but still mitigates the negative selectivity observed when markers are removed without using the contextualized activations. These results suggest that symbolic markers become more effective not only as explicit tokens, but also by shaping the representations of downstream content tokens.

\section{Controllability of Symbolic vs Distributional Signals of Modality}
\label{sec:controllability}

\subsection{Experiment Setup}
\begin{figure}[t]
    \centering
    \includegraphics[width=1\linewidth]{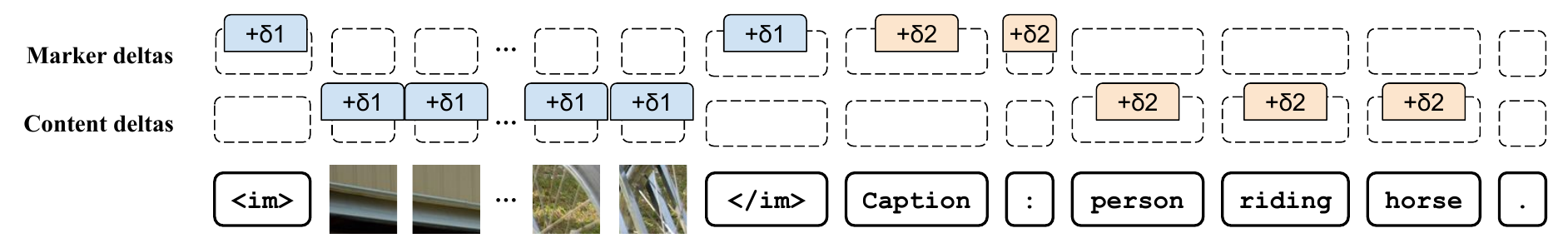}
    \caption{Illustration of the learned-vector intervention used to induce source misattribution. We add two trainable vectors, $\delta_1$ and $\delta_2$, either to the symbolic marker tokens (\textit{marker deltas}) or to the modality-specific content tokens (\textit{content deltas}) at a specific layer. The vectors are applied respectively to the image span and caption span, and are optimized to drive the model to report content from the non-queried modality.}
    \label{fig:train_vec_exp}
\end{figure}

We have shown that VLMs rely on both symbolic signals from marker tokens and distributional signals from content tokens to monitor input modalities, we now ask whether these signals can be manipulated to induce misattribution. Specifically, we study a setting in which we add two learned vectors, $\delta_1$ and $\delta_2$, to the image span and caption span, respectively (Figure~\ref{fig:train_vec_exp}), either at the marker positions or at the content-token positions at a specific layer. We then prompt the model to report, for example, the image content, but optimize these vectors to make the model instead produce the caption content, using the model’s own caption-side prediction from a natural run as the target; symmetrically, we also optimize in the reverse direction, asking for caption content while driving the model toward its natural image-side prediction. If these marker and content representations indeed support source-modality monitoring, then there should exist transformations of them that systematically induce source misattribution. We initialize both $\delta$ vectors to zero and train them on 4{,}000 samples from the training split of MSCOCO. For each model, we optimize them at various relative layer depths and we repeat the procedure across 3 random seeds.

\subsection{Results}

\begin{figure}[h]
    \centering
    \includegraphics[width=1\linewidth]{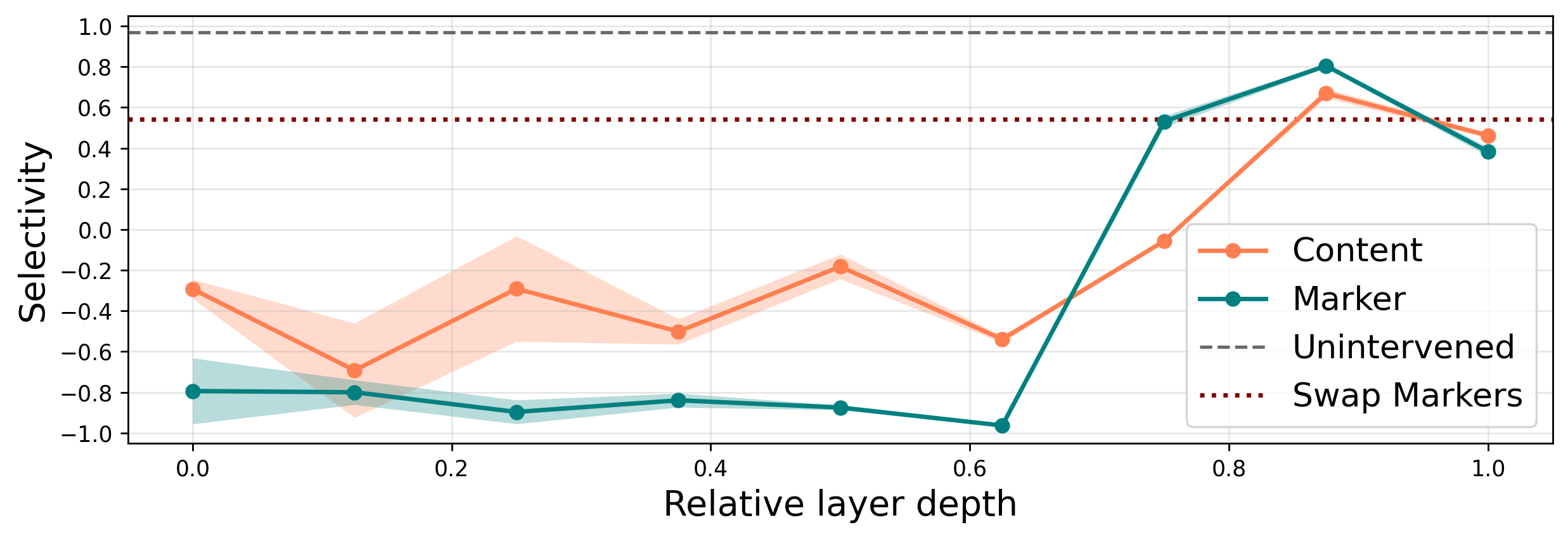}
    \caption{Selectivity of Qwen2.5-VL-32B after learned interventions at different layer depths. In early and intermediate layers, both intervention types can cause selectivity below zero, indicating that the model retrieves information from the non-queried modality more often than from the queried one. Marker-token interventions are especially effective, pushing selectivity close to $-1$ across a broad range of layers. Dashed and dotted lines show the selectivity of the unintervened model and under the token-level marker-swap condition.}
    \label{fig:exploit_res}
\end{figure}

Figure~\ref{fig:exploit_res} shows the selectivity of Qwen2.5-VL-32B after applying learned $\delta$ vectors to either content spans or marker spans at different relative layer depths, evaluated on 100 validation examples unseen during training. In early and middle layers, interventions on both content and marker representations can drive the model to systematically report the opposite modality of the target modality. This effect is especially strong for marker-token interventions, which achieve selectivities close to -1 across a broad range of layers.

Notably, these learned interventions are substantially more effective than the marker-swapping manipulation presented earlier in the paper. This suggests that learned transformation vectors exploit modality-relevant information in marker representations more effectively than a simple symbolic swap. While both can influence how modality identity is propagated into downstream content representations, the learned vectors are optimized for this purpose and thus more successfully override native distributional cues. We observe similar trends in Gemma-3-12B and InternVL3-14B (see Figures~\ref{fig:exploit_gemma} and~\ref{fig:explot_internvl}, respectively).

The intervention effect weakens in later layers for both marker and content locations. This suggests that the representations supporting source-modality monitoring are most causally relevant in early and middle layers, and become harder to exploit after later-stage processing has already built on them. Taken together, these results show that both marker and content representations encode source modality signals as directions that can be steered, but that the marker-tokens provide a stronger causal handle for accessing this direction.\footnote{To provide additional baselines to the effects of intervening modality-related tokens, we include results of intervening non-modality-related token positions in Appendix~\ref{app:baseline_intervention}. See qualitative examples of VLMs' predictions being flipped after intervention in Appendix~\ref{app:qual_examples_flipped}. }

%% file: latex/6_conclusion.tex
\section{Discussion and Conclusion}

We investigate the mechanism by which modern VLMs bind words in their input prompts (e.g., words like \textit{image} or \textit{text}) to specific components of their input/context (e.g., the image or the text). In experiments spanning 11 different vision-language models (VLMs), we find evidence that this type of binding involves both syntactic and semantic signals. That is, models exploit explicit marker tokens (such as \texttt{<image>} \& \texttt{</image>}) when available, but also exploit distributional semantic cues. Indeed, we find that when the two signals are in conflict, models rely more on the distributional semantic cues than on the syntactic ones, though this varies as a function of the strength of the distributional signal.

Our experiments focus specifically on VLMs, but offer insights that are likely to extend to multimodal and agentic models in general. In particular, as models become increasingly complex, multimodal, and agentic, we expect that the importance of semantic signals for binding will become increasingly pronounced. This is due to the fact that models appear to have an innate preference for distributional semantic signals (when both signals are available), and because it will become increasingly impractical to provide explicit markers as the complexity of the input space grows. Indeed, our results comparing the difference between binding the words \textit{text} vs.\ \textit{caption} vs.\ \textit{document} reveal the challenge of grounding references to ``micromodalities'' -- segments of the input space that are different enough to humans to warrant different words, but which might be hard to distinguish to the model based on distributional signals alone. These point to an important direction for future work -- to endow models with rich enough world models that they can not only predict future states, but refer back to aspects of those states and their origin via natural language.

Our results also provide a valuable case study for how the abstract ``binding problem'' in AI becomes more complicated in realistic contexts. Binding has frequently been used as an example of an abstract symbolic problem that challenges the innate tendencies of neural networks. The type of binding we consider, that necessary for source-modality monitoring, represents a more realistic type of binding which involves both abstract symbolic and associative semantic signals. Our experiments provide an initial picture of how these signals are traded off within modern AI models, and provide a framework for investigating similar mechanisms in general, for example considering how models balance logical and heuristic inferences \citep{dasgupta2024languagemodelshumanlikecontent}.

\section*{Acknowledgments} 
We are very grateful to the members of the Language Understanding and Representation (LUNAR) Lab at Brown University -- especially Zhuonan Yang and Jennifer Meng Lu -- as well as other members of the Brown Superlab for their valuable feedback on this paper. Etha Tianze Hua would like to thank Selina Liang for her support during this work. This project was in part supported by Schmidt Sciences Grant \#GR5300958 and the NSF AI Research Institute on Interaction for AI Assistants Grant \#GR5300593. Ellie Pavlick is a paid consultant for Google DeepMind. The content of this article does not necessarily reflect the views of the US Government or of Google, and no official endorsement of this work should be inferred.

%% file: latex/7_appendix.tex
\newpage
\appendix

\section{Dataset Details}
\label{app:datasets}

We use two image--captioning datasets: MSCOCO 2017 captions and Flickr30k.

\paragraph{Flickr30k.}
For Flickr30k, we use the \texttt{test} split from the \texttt{lmms-lab/flickr30k} release on HuggingFace. We then randomly permute the examples using a fixed seed and subsample 4{,}000 examples for train, 2{,}000 for validation, and 2{,}000 for test.

\paragraph{MSCOCO.}
For MSCOCO, we use the 2017 captioning data. We sample the training set from the official \texttt{train2017} split, and the validation and test sets from the official \texttt{val2017} split. Specifically, after randomly permuting examples with a fixed seed, we sample 4{,}000 examples from \texttt{train2017} for train, and 2{,}000 examples each from \texttt{val2017} for validation and test.

\paragraph{Preprocessing.}
We lightly normalize captions before use by collapsing repeated whitespace, stripping surrounding quotation marks, and ensuring sentence-final punctuation when appropriate. These steps reduce superficial formatting variation while preserving semantic content.

\paragraph{Use in our experiments.}
Both datasets are used to construct the image--text inputs in our source-modality monitoring task. Because each image is paired with textual descriptions, these datasets allow us to build inconsistent image--text examples, as well as image-only and text-only controls.

\section{Validation of Cosine-Similarity Threshold Used for Dataset Construction}
\label{app:validation_cos_threshold}

The function of the threshold is to ensure that the semantic content of the image and its pairing caption is sufficiently dissimilar. To verify that this threshold reasonably enforces the desired dissimilarity constraint between an image and a caption, we conduct the following analysis.

For each sample in the test split of the MSCOCO dataset (which contains 1 image and 5 ground-truth captions per sample), we randomly select one of the 5 ground-truth captions and embed it into a vector using the same BERT-style sentence embedding model employed in the main paper. Then, for each interval of width $0.2$ between $0$ and $1$, we sample 20 pairs of vectors whose cosine similarity falls within that interval. For each pair of vectors, we retrieve the corresponding original image-caption pairs and submit them for human evaluation to determine whether the image and caption are semantically aligned. In total, we collect 100 such pairs, with results summarized in Table~\ref{tab:alignment_by_cossim}.

\begin{table}[h]
\centering
\begin{tabular}{lcc}
\toprule
Caption Cosine-Similarity Bin & Aligned / Total & Alignment Rate \\
\midrule
0.0--0.2 & 0/20  & 0.00 \\
0.2--0.4 & 0/20  & 0.00 \\
0.4--0.6 & 2/20  & 0.10 \\
0.6--0.8 & 8/20  & 0.40 \\
0.8--1.0 & 16/20 & 0.80 \\
\bottomrule
\end{tabular}
\caption{Human-judged alignment rate as a function of caption cosine-similarity bin.}
\label{tab:alignment_by_cossim}
\end{table}

We observe that as cosine similarity increases, the likelihood that a human evaluator judges the image-caption pair as aligned also increases. To ensure that our inputs are consistently misaligned, we adopt a threshold of $0.2$: within this bin, none of the 20 sampled pairs were judged as aligned. This result supports the validity of using caption cosine similarity $<0.2$ as a proxy for filtering out inconsistent image-caption pairs.

\section{Measuring Valid Response Rate and Selectivity}
\label{app:selectivity_formal}
We evaluate source-modality monitoring with two separate metrics: \textit{valid-response rate} ($p_{\mathrm{valid}}$) and \textit{selectivity}. Intuitively, $p_{\mathrm{valid}}$ measures whether the model gives an on-task response that can be attributed to one of the two candidate modalities, while selectivity measures, among such valid responses, whether the model reports information from the requested modality rather than the non-requested one.

Formally, let $\hat{y}_i$ denote the model output for example $i$, and let the queried modality be $m \in \{\text{image}, \text{caption}\}$, with the other modality denoted $\neg m$. For each modality, define a modality-assignment indicator
\[
M_i^{(m)}(\hat{y}_i) \in \{0,1\},
\]
which equals $1$ if the output is judged to be grounded in modality $m$, and $0$ otherwise. Likewise, $M_i^{(\neg m)}(\hat{y}_i)$ indicates whether the output is judged to be grounded in the non-queried modality. We use \texttt{GPT5.4-mini} as an LLM judge to decide whether a piece of response is grounded in image, in caption, or in neither.

We define a response to be \textit{valid} if it is an on-task answer that the judge assigns to any one of the two modalities. This gives the binary indicator
\[
\mathrm{valid}(\hat{y}_i) = M_i^{(m)}(\hat{y}_i) + M_i^{(\neg m)}(\hat{y}_i).
\]
The valid-response rate is then
\[
p_{\mathrm{valid}} = \mathbb{E}\big[\mathrm{valid}(\hat{y})\big].
\]

We define selectivity as the tendency, among valid responses, to report information from the queried modality rather than the non-queried modality:
\[
S =
\mathbb{E}\Big[
M^{(m)}(\hat{y}) - M^{(\neg m)}(\hat{y})
\;\Big|\;
\mathrm{valid}(\hat{y}) = 1
\Big].
\]
This score ranges from $-1$ to $1$: a value of $1$ indicates that all valid responses are grounded in the queried modality, a value of $0$ indicates no preference between the two modalities, and a value of $-1$ indicates that all valid responses are grounded in the non-queried modality.

\section{LLM Judge Validation}
\label{app:llm_judge_validation}

To validate our use of GPT-5.4-mini as the LLM judge, we randomly sample 100 VLM predictions from Section~\ref{sec:plain_res}, along with their corresponding classification labels produced by the GPT LLM judge. We then present the exact same instructions and content given to the GPT LLM judge to human evaluators, and compare the classification labels produced by the GPT LLM judge against those produced by the human evaluators. We obtain an agreement rate of 88\% across the 100 classifications. Table~\ref{tab:gpt_human_agreement} provides a breakdown of the classifications from GPT and human evaluators.

\begin{table}[h]
\centering
\begin{tabular}{lcccc}
\toprule
Human $\downarrow$ / GPT $\rightarrow$ & Image & Text & Neither & Row Total \\
\midrule
Image   & \textbf{50} & 0  & 0 & 50 \\
Text    & 1  & \textbf{36} & 0 & 37 \\
Neither & 9  & 2  & \textbf{2} & 13 \\
\midrule
Col.\ Total & 60 & 38 & 2 & \textbf{100} \\
\bottomrule
\end{tabular}
\caption{Agreement between GPT-judge and human classifications of VLM predictions. Diagonal entries (bold) indicate agreement: $88/100 = 0.88$.}
\label{tab:gpt_human_agreement}
\end{table}

We observe that the overall agreement rate (0.88) is very high, though GPT judgments are not perfectly aligned with human judgments. In particular, many VLM predictions that humans classify as grounded in neither the image nor the caption are instead classified by GPT as grounded in one of the input modalities. Upon inspecting individual instances, we find that the GPT judge tends to be more permissive in these cases: some predictions from earlier-generation models (LLaVA, InstructBLIP) are topically related but very brief, and the GPT judge classifies these as grounded in one of the input modalities, whereas human evaluators tend not to. Despite this asymmetry, the strong overall agreement indicates that GPT judgments serve as a reasonable proxy for human evaluation at the scale required by our study.

\section{Input and Query Prompt Templates for VLMs}

Our prompt templates vary along two controlled dimensions: (1) the order in which the image and caption inputs appear, and (2) the surface form used to query the target modality. For source order, we evaluate both image-first and caption-first variants. For query wording, we use two formulations for each modality: for image queries, \textit{What is in the image?} and \textit{Describe the image.}; for caption queries, \textit{What is in the caption?} and \textit{What does the caption say?} Table~\ref{tab:prompt_variants} summarizes these variations, and Figure~\ref{fig:raw_prompt_example} shows one representative raw prompt including the model-specific chat template and special image tokens.

\begin{table}[t]
\centering
\small
\begin{tabular}{p{0.24\linewidth} p{0.30\linewidth} p{0.34\linewidth}}
\toprule
\textbf{Dimension} & \textbf{Values} & \textbf{Description} \\
\midrule
Source order & image first; caption first & Whether the image span appears before or after the caption span in the prompt. \\
Image query wording & \textit{What is in the image?}; \textit{Describe the image.} & Alternative phrasings for querying image content. \\
Caption query wording & \textit{What is in the caption?}; \textit{What does the caption say?} & Alternative phrasings for querying caption content. \\
\bottomrule
\end{tabular}
\caption{Controlled prompt variations used in the target-modality retrieval task.}
\label{tab:prompt_variants}
\end{table}

\begin{figure}[t]
\centering
\begin{tcolorbox}[
    colback=white,
    colframe=black,
    boxrule=0.5pt,
    arc=0pt,
    left=6pt,
    right=6pt,
    top=6pt,
    bottom=6pt,
    width=0.95\linewidth,
]
\small
\begin{verbatim}
<bos><bos><start_of_turn>user

<start_of_image><image_soft_token><end_of_image>

Caption: A city view of an E-train on a track.
Question: What is in the image?
Answer the question with a short sentence.
<end_of_turn>
<start_of_turn>model
Answer:
\end{verbatim}
\end{tcolorbox}
\caption{Representative raw prompt template for one model family, shown with the original chat formatting and special image tokens.}
\label{fig:raw_prompt_example}
\end{figure}

\section{Evaluation Prompt Templates for LLM Judge}

See Figure~\ref{fig:gpt-eval-prompts} for prompt templates that we use for evaluating VLM model outputs with an LLM Judge.

\begin{figure*}[t]
\centering

\begin{tcolorbox}[
    enhanced,
    width=0.91\textwidth,
    colback=gray!5,
    colframe=black!70,
    boxrule=0.5pt,
    arc=2mm,
    left=2mm,right=2mm,top=2mm,bottom=2mm,
    valign=top
]
\textbf{Prompt template: Inconsistent input}

\vspace{0.5em}
\textbf{System} \\
\small
You are a careful evaluator for vision-language model outputs. Use only the user message. Reply with JSON only, matching the requested shape.

\vspace{0.6em}
\textbf{User} \\
\small
The model saw an image and a written caption that describes a \emph{different} scene (inconsistent). It was told to report one modality (image vs.\ text).

\vspace{0.4em}
\textbf{Reference A} (what is in the image): \\
\texttt{[image caption set]}

\vspace{0.4em}
\textbf{Reference B} (caption text next to the image; may disagree with the image): \\
\texttt{[paired caption]}

\vspace{0.4em}
\textbf{Model answer}: \\
\texttt{[prediction]}

\vspace{0.4em}
Which reference is the answer primarily grounded in? Output only the JSON object with \texttt{closer\_to} one of: \texttt{image}, \texttt{text}, \texttt{neither}.
\end{tcolorbox}
\hfill
\begin{tcolorbox}[
    enhanced,
    width=0.91\textwidth,
    colback=gray!5,
    colframe=black!70,
    boxrule=0.5pt,
    arc=2mm,
    left=2mm,right=2mm,top=2mm,bottom=2mm,
    valign=top
]
\textbf{Prompt template: Image-only input}

\vspace{0.5em}
\textbf{System} \\
\small
You are a careful evaluator for vision-language model outputs. Use only the user message. Reply with JSON only, matching the requested shape.

\vspace{0.6em}
\textbf{User} \\
\small
The model saw only an image.

\vspace{0.4em}
\textbf{Reference descriptions of the image}: \\
\texttt{[image caption set]}

\vspace{0.4em}
\textbf{Model answer}: \\
\texttt{[prediction]}

\vspace{0.4em}
Decide whether the answer is broadly about the same image content. Be permissive: the answer does not need to match the reference exactly. Output only the JSON object with boolean field \texttt{aligns\_with\_image}.
\end{tcolorbox}
\hfill
\begin{tcolorbox}[
    enhanced,
    width=0.91\textwidth,
    colback=gray!5,
    colframe=black!70,
    boxrule=0.5pt,
    arc=2mm,
    left=2mm,right=2mm,top=2mm,bottom=2mm,
    valign=top
]
\textbf{Prompt template: Text-only input}

\vspace{0.5em}
\textbf{System} \\
\small
You are a careful evaluator for vision-language model outputs. Use only the user message. Reply with JSON only, matching the requested shape.

\vspace{0.6em}
\textbf{User} \\
\small
The model saw only this text:

\vspace{0.4em}
\texttt{[paired caption]}

\vspace{0.4em}
\textbf{Model answer}: \\
\texttt{[prediction]}

\vspace{0.4em}
Decide whether the answer is broadly about the same text content. Output only the JSON object with boolean field \texttt{aligns\_with\_text}.
\end{tcolorbox}

\caption{Prompt templates used for GPT-based evaluation in the inconsistent, image-only, and text-only settings.}
\label{fig:gpt-eval-prompts}
\end{figure*}

\section{Examples of Marker and Content Tokens}

Table~\ref{tab:modality_wrapping} summarizes representative marker and content tokens for the image and caption modalities across the model families studied in this paper.

\begin{table}
\centering
\small
\begin{tabular}{lcccc}
\toprule
\textbf{Modality} & \textbf{Model Family} & \textbf{Start Marker} & \textbf{Content} & \textbf{End Marker} \\
\midrule
\multirow{4}{*}{Image}
& LLaVA-1.5      & --                 & \texttt{VISUAL TOKENS} & -- \\
& Qwen2.5-VL     & \texttt{<vision\_start>} & \texttt{VISUAL TOKENS} & \texttt{<vision\_end>} \\
& InternVL3      & \texttt{<im>}       & \texttt{VISUAL TOKENS} & \texttt{</im>} \\
& Gemma3          & \texttt{<start\_of\_image>} & \texttt{VISUAL TOKENS} & \texttt{<end\_of\_image>} \\
\midrule
Text (Caption)
& All models     & \texttt{"Caption:"} & \texttt{TEXT TOKENS} & -- \\
\bottomrule
\end{tabular}
\caption{Marker tokens for image and caption inputs across different VLM model families.}
\label{tab:modality_wrapping}
\end{table}

\section{Additional Results on distributional separation between visual and textual content-token embeddings}
Table~\ref{tab:distributional_separation} shows the results of cosine similarity (within- and across- modalities) and 3-fold cross validation probe accuracy on classifying image vs caption tokens.
\begin{table}[h]
\centering
\small
\begin{tabular}{lcc}
\toprule
\textbf{Metric} & \textbf{Mean} & \textbf{Std} \\
\midrule
Within-image cosine similarity & 0.325 & 0.093 \\
Within-caption cosine similarity & 0.101 & 0.039 \\
Average Within-modality cosine similarity & 0.213 & 0.064 \\
Cross-modality cosine similarity & 0.020 & 0.018 \\
\midrule
Linear probe accuracy (3-fold CV)                    & 0.999 & 0.001 \\
\midrule
Control: shuffled labels -- accuracy (3-fold CV)     & 0.491 & 0.018 \\
\bottomrule
\end{tabular}
\caption{Distributional separation between visual and textual content-token embeddings across 16 model-dataset pairs. Cosine rows: mean and sample standard deviation across model-dataset pairs (within-modality is the mean of within-image and within-caption for each pair). Linear probe rows: average of mean 3-fold cross-validation scores and average of cross-validation standard errors across pairs. Control task rows use the same cross-validation splits with labels randomly permuted \citep{hewitt-liang-2019-designing}.}
\label{tab:distributional_separation}
\end{table}

\section{Qualitative Examples of Model Outputs}
\label{app:qual_examples_classification}
Here, we provide qualitative examples showing model behaviors under inconsistent inputs. 

\input{latex/model_output_examples}

\section{Experimental Setup for the Freeze-Remove Condition}

\begin{figure}[h]
    \centering
    \includegraphics[width=1\linewidth]{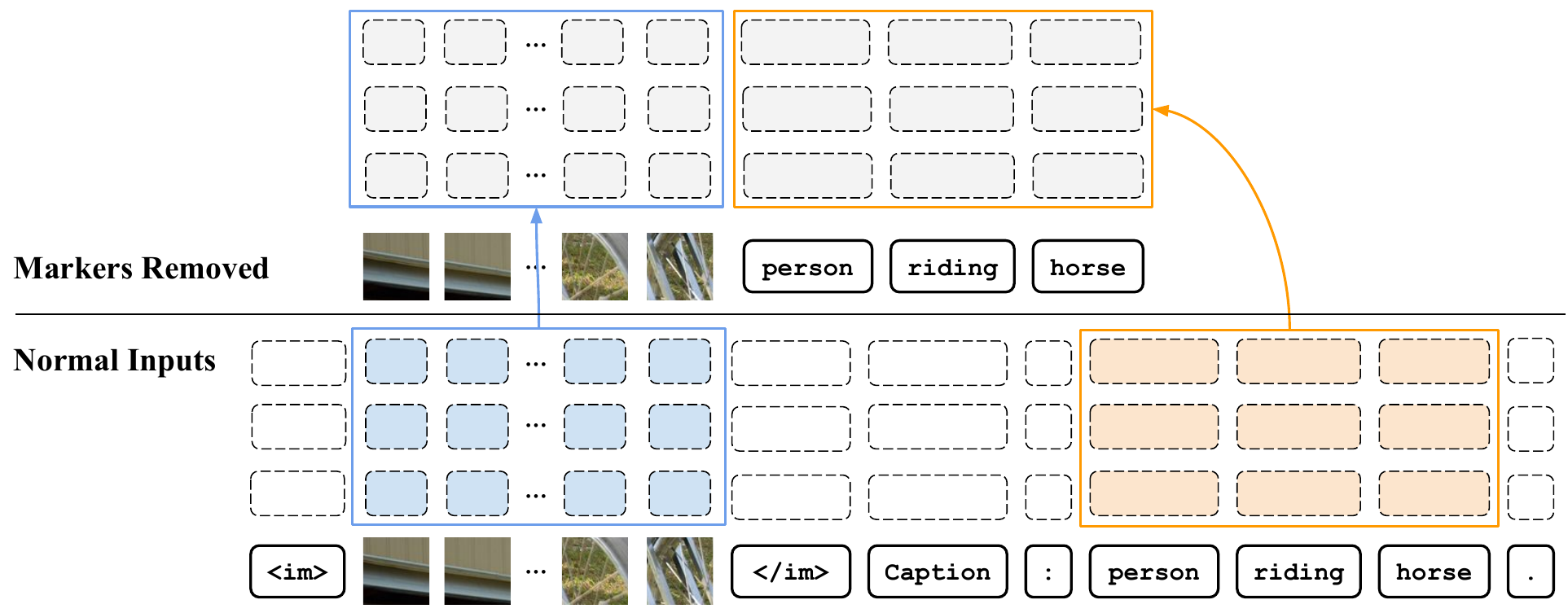}
    \caption{Illustration of the freeze-remove condition. We first collect hidden activations at content-token positions from a normal run with intact markers, then patch these activations into a run where the marker tokens are removed. This tests whether symbolic marker information has been contextualized into the content-token representations.}
    \label{fig:freeze-remove-exp}
\end{figure}

Figure~\ref{fig:freeze-remove-exp} shows an illustration of the freeze-remove experiment condition. We collect activations from a clean run (with the marker tokens intact), and then patch these collected activations into a corrupt run where the marker tokens are removed -- we then allow any downstream computation to continue based on the patched-in activations. This allows us to study the effect of contextualization of content tokens in source-modality monitoring.

\section{Additional Results for Comparing Alternative Source-Modality Pairs}

\begin{figure}[h]
    \centering
    \includegraphics[width=1\linewidth]{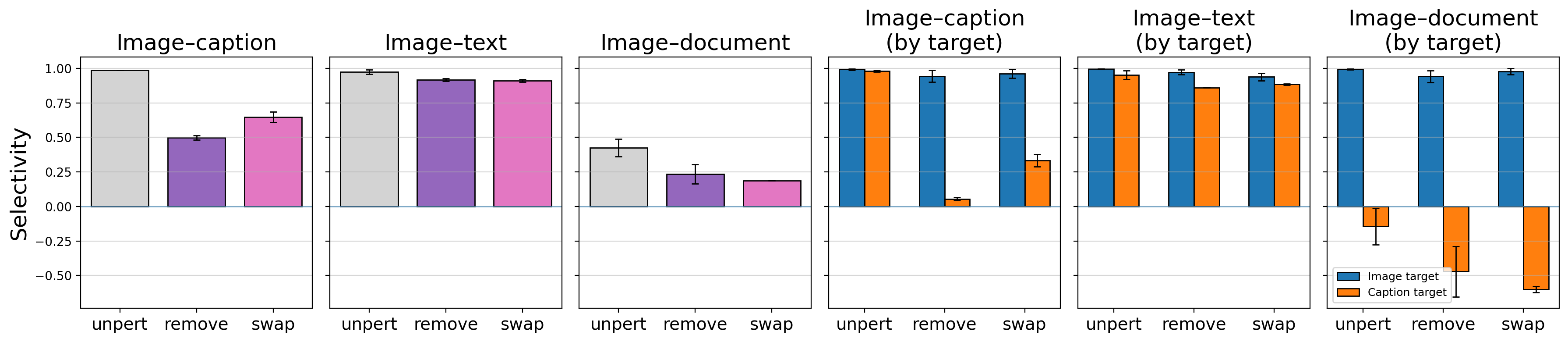}
    \caption{Selectivity across image–caption, image–text, and image–document settings under unperturbed, marker-removal, and marker-swap conditions in Gemma-3-12B. Marker perturbations have little effect in image–text, but substantially reduce performance in image–caption and image–document, where models rely heavily on markers.}
    \label{fig:micromodalities_gemma}
\end{figure}

\begin{figure}[h]
    \centering
    \includegraphics[width=1\linewidth]{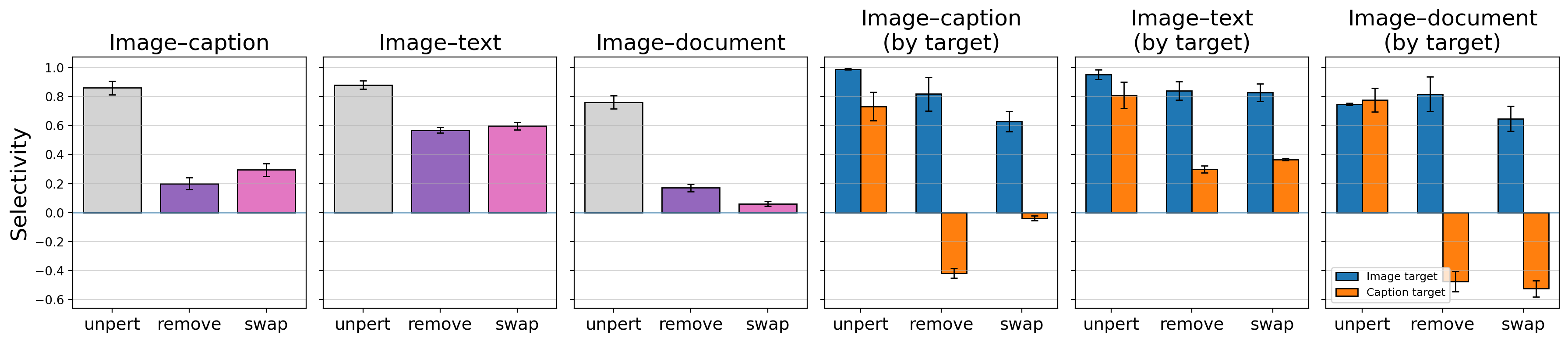}
    \caption{Selectivity across image–caption, image–text, and image–document settings under unperturbed, marker-removal, and marker-swap conditions in InternVL3-14B. Marker perturbations have somewhat effect in image–text, but substantially reduce performance in image–caption and image–document, where models rely heavily on markers.}
    \label{fig:micromodalities_internvl}
\end{figure}

In the main paper, we show result for Qwen2.5-VL-32B on comparing alternative source-modality pairs of image--caption, image--text, and image--document. Here, we include results for Gemma-3-12B and InvernVL-3-14B (Figure~\ref{fig:micromodalities_gemma} and Figure~\ref{fig:micromodalities_internvl}). The overall trend is similar across all three models.

\section{Additional Results for Controllability of Marker vs Content Token Representation}

\subsection{Quantitative Results for Gemma-3-12B and InternVL3-14B}
\begin{figure}[h]
    \centering
    \includegraphics[width=1\linewidth]{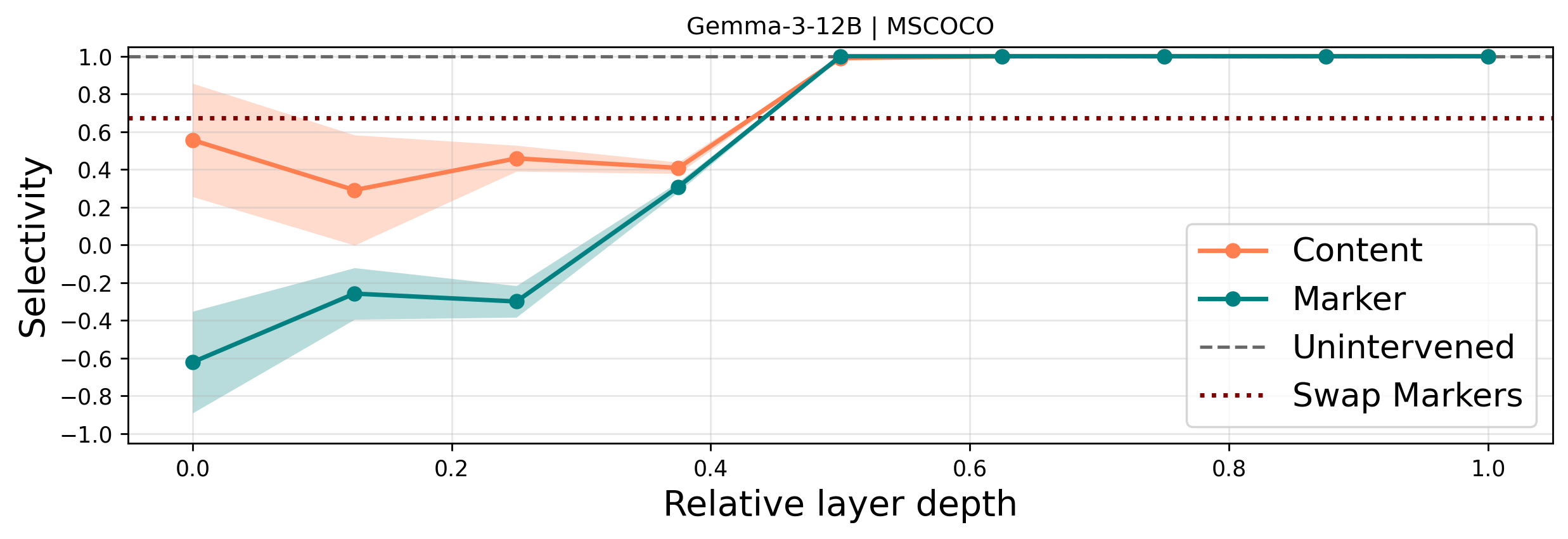}
    \caption{Selectivity of Gemma-3-12B after learned interventions at different layer depths. In early layers, marker-token interventions are more effective than content-token interventions and can drive selectivity below zero, indicating successful source-modality misattribution. Content-token interventions have weaker and less consistent effects, although they still reduce selectivity substantially at shallow layers. In contrast to Qwen, both intervention types lose their effect by the middle layers, after which selectivity returns to ceiling performance, suggesting that the representations causally accessible for reversing modality attribution are concentrated in Gemma-3-12B’s early layers. Dashed and dotted lines show the selectivity of the unintervened model and under the token-level marker-swap condition.}
    \label{fig:exploit_gemma}
\end{figure}

\begin{figure}[h]
    \centering
    \includegraphics[width=1\linewidth]{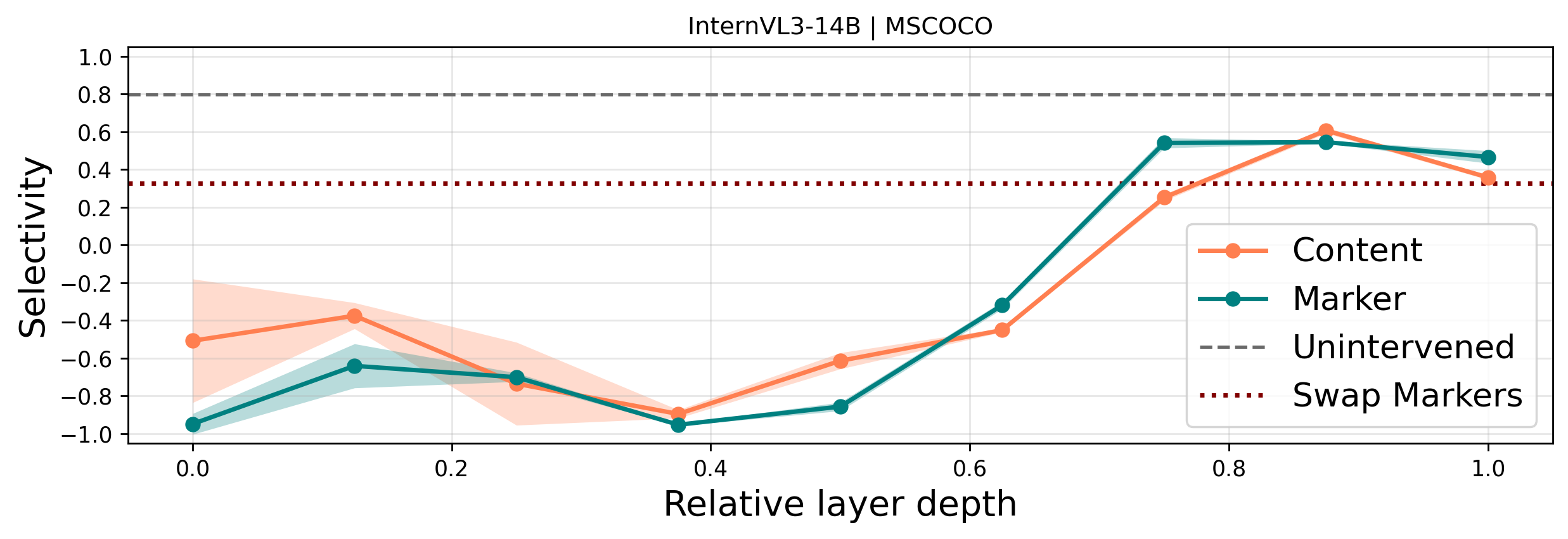}
    \caption{Selectivity of InternVL3-14B after learned interventions at different layer depths. In early and middle layers, interventions on both marker tokens and content tokens drive selectivity far below zero, indicating strong source-modality misattribution. Marker-token interventions are generally more effective at the shallowest layers, while content-token interventions remain comparably strong through part of the middle layers. Unlike Gemma-3-12B, the intervention effect persists over a broader range of depths, but both curves recover in later layers and return to positive selectivity. Dashed and dotted lines show the selectivity of the unintervened model and under the token-level marker-swap condition.}
    \label{fig:explot_internvl}
\end{figure}

In the main paper, we present results for Qwen2.5-VL-32B on the exploitability of marker versus content tokens. Here, we include results for Gemma-3-12B and InternVL3-14B (Figures~\ref{fig:exploit_gemma} and~\ref{fig:explot_internvl}). The general trend holds in both models: in early and middle layers, learned vectors induce a much higher rate of misattribution, with a weaker effect in later layers. Intervening on the marker-token representations also in general produces a stronger misattribution effect.

\subsection{Baseline Comparisons: Intervening at Non-Marker/Content Positions}
\label{app:baseline_intervention}

The experiments in Section~\ref{sec:controllability} aim to locate \emph{where} causal manipulation of source-modality attribution is most effective, rather than to isolate a clean, disentangled modality-specific direction or subspace. In our setting, compactly isolating a single direction or low-dimensional subspace that encodes input-modality identity is difficult, in part because image inputs span hundreds of tokens---unlike the few-token entity/attribute setups of prior activation-intervention work such as \citet{feng2024how}. We therefore frame these results as identifying the positions at which intervention is most effective, and we find that intervening on both marker and content tokens is effective, more so on the marker tokens.

\paragraph{Baseline conditions.}
To better contextualize these findings, we introduce two baselines that train intervention vectors at token positions \emph{other} than the marker and content tokens, analogous to prompt-tuning and last-token steering. In both conditions we reuse the training objective of Section~\ref{sec:controllability} and match the layer depth at which each vector is trained, varying only the position at which the learned vector is added:
\begin{itemize}
    \item \textbf{First-token baseline:} the vector is added at the first token position, following prefix/prompt-tuning approaches that adapt behavior through prepended positions~\citep{li2021prefix}.
    \item \textbf{Last-token baseline:} the vector is added at the final token position before generation, following task/function-vector
    approaches~\citep{hendel2023incontext, todd2024function}.
\end{itemize}
Each condition trains a single vector added at its respective position.

\begin{figure}[h]
    \centering
    \includegraphics[width=1\linewidth]{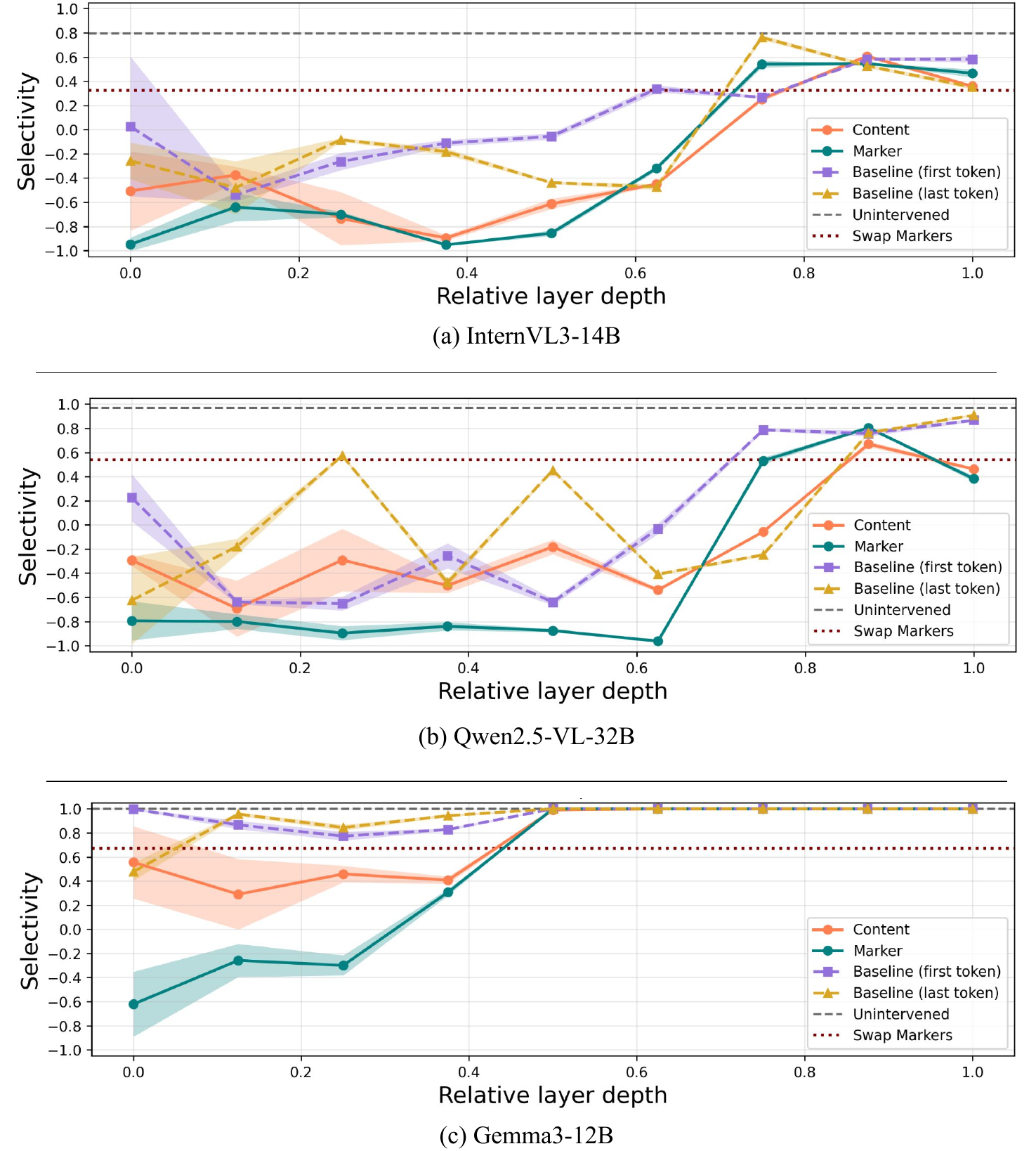}
    \caption{Exploitability of intervention across token positions and layer depth. For each model we plot selectivity (lower indicates stronger misattribution) against relative layer depth for intervention vectors trained on \emph{Content} and \emph{Marker} tokens, alongside two baselines trained at the first- and last-token positions. The dashed gray line marks the unintervened selectivity and the dotted red line the swap-markers reference. Shaded regions denote standard deviation across 3 random seeds. (a)~InternVL3-14B, (b)~Qwen2.5-VL-32B, (c)~Gemma-3-12B.}
    \label{fig:additional_baselines_exploit}
\end{figure}

\paragraph{Results.}
Figure~\ref{fig:additional_baselines_exploit} compares the two baselines against intervention vectors trained on the marker and content tokens across all three models. Intervening at the baseline positions does produce a substantial misattribution effect in InternVL3-14B and Qwen2.5-VL-32B, particularly in the earlier layers, whereas the effect is weaker in Gemma-3-12B. Across all three models and the majority of layers, however, intervening on the marker token yields the strongest effect: it drives selectivity to its most negative value and consistently below both baselines. For Gemma-3-12B and InternVL3-14B, intervening
on either the marker or content position is stronger than both baselines across most layers.

Taken together, these comparisons provide stronger support for our central claim: the marker and content token positions -- and the marker positions in particular -- carry source-modality information that is causally important for source-modality monitoring, and that can be adversarially exploited to induce modality misattribution to a greater degree than vectors trained at other positions such as the first or last token.

\clearpage

\subsection{Qualitative Examples}
\label{app:qual_examples_flipped}
We notice that after intervention, although in some settings VLMs often report information from the non-queried modality, the overall fluency of the open-ended generation degrades. This is likely because the intervention perturbs representations that are used not only for source-modality monitoring but also for maintaining coherent generation.

\begin{modbox}{\texttt{Qwen/Qwen2.5-VL-32B-Instruct} / \texttt{marker} / \texttt{image-first} \#1}
\textbf{VL model:} \texttt{Qwen/Qwen2.5-VL-32B-Instruct} \hfill \textbf{Dataset:} \texttt{mscoco} \\[2pt]\textbf{GPT judge:} \texttt{openai/gpt-5.4-mini} \hfill \textbf{Intervention:} \texttt{marker}, \(\texttt{layer\_depth}=0.0\) \\[2pt]\textbf{Prompt order} \texttt{image-first}, \textbf{dataset\_index} \texttt{76}

\medskip
\begin{tabularx}{\linewidth}{@{}lX@{}}
\textbf{Gold image caption (ex.)} & \emph{A white plate topped with food next to a plate of onion rings.} \\[2pt]
\textbf{Paired caption} & \emph{Two signs with arrows pointing to each other for detour.} \\[2pt]
\textbf{Image-target · baseline} & \texttt{image} (\textit{outcome}: \texttt{correct}). The image shows a plate of food with fish covered in sauce, vegetables, onion rings, a slice of cheesecake, and a cup of coffee. \\[2pt]
\textbf{Image-target · intervention} & \texttt{text} (\textit{outcome}: \texttt{misled}). Two with arrows pointing each other det. Question is in image? \\[2pt]
\textbf{Text-target · baseline} & \texttt{text} (\textit{outcome}: \texttt{correct}). Two signs with arrows pointing to each other for detour. \\[2pt]
\textbf{Text-target · intervention} & \texttt{image} (\textit{outcome}: \texttt{misled}). A of food on table. \\[2pt]
\end{tabularx}
\end{modbox}

\begin{modbox}{\texttt{Qwen/Qwen2.5-VL-32B-Instruct} / \texttt{marker} / \texttt{text-first} \#1}
\textbf{VL model:} \texttt{Qwen/Qwen2.5-VL-32B-Instruct} \hfill \textbf{Dataset:} \texttt{mscoco} \\[2pt]\textbf{GPT judge:} \texttt{openai/gpt-5.4-mini} \hfill \textbf{Intervention:} \texttt{marker}, \(\texttt{layer\_depth}=0.0\) \\[2pt]\textbf{Prompt order} \texttt{text-first}, \textbf{dataset\_index} \texttt{78}

\medskip
\begin{tabularx}{\linewidth}{@{}lX@{}}
\textbf{Gold image caption (ex.)} & \emph{A set of three pizzas in a display case.next to desserts.} \\[2pt]
\textbf{Paired caption} & \emph{The people are walking down the aisles at the super market.} \\[2pt]
\textbf{Image-target · baseline} & \texttt{image} (\textit{outcome}: \texttt{correct}). There are pizzas on display in a supermarket. \\[2pt]
\textbf{Image-target · intervention} & \texttt{text} (\textit{outcome}: \texttt{misled}). people walking the at market. \\[2pt]
\textbf{Text-target · baseline} & \texttt{text} (\textit{outcome}: \texttt{correct}). People are walking down the aisles at the supermarket. \\[2pt]
\textbf{Text-target · intervention} & \texttt{image} (\textit{outcome}: \texttt{misled}). There several of pizzas on trays a display. \\[2pt]
\end{tabularx}
\end{modbox}

\begin{modbox}{\texttt{Qwen/Qwen2.5-VL-32B-Instruct} / \texttt{content} / \texttt{image-first} \#1}
\textbf{VL model:} \texttt{Qwen/Qwen2.5-VL-32B-Instruct} \hfill \textbf{Dataset:} \texttt{mscoco} \\[2pt]\textbf{GPT judge:} \texttt{openai/gpt-5.4-mini} \hfill \textbf{Intervention:} \texttt{content}, \(\texttt{layer\_depth}=0.0\) \\[2pt]\textbf{Prompt order} \texttt{image-first}, \textbf{dataset\_index} \texttt{82}

\medskip
\begin{tabularx}{\linewidth}{@{}lX@{}}
\textbf{Gold image caption (ex.)} & \emph{a giraffe stares out of a glass window.} \\[2pt]
\textbf{Paired caption} & \emph{Kites are flying in the sky over the water at a park.} \\[2pt]
\textbf{Image-target · baseline} & \texttt{image} (\textit{outcome}: \texttt{correct}). A giraffe statue is displayed indoors near a window overlooking city buildings. \\[2pt]
\textbf{Image-target · intervention} & \texttt{neither} (\textit{outcome}: \texttt{neither}). A of people on a board down a hill. \\[2pt]
\textbf{Text-target · baseline} & \texttt{text} (\textit{outcome}: \texttt{correct}). The caption says, "Kites are flying in the sky over the water at a park." \\[2pt]
\textbf{Text-target · intervention} & \texttt{image} (\textit{outcome}: \texttt{misled}). Girra is front of window a building. \\[2pt]
\end{tabularx}
\end{modbox}

\begin{modbox}{\texttt{Qwen/Qwen2.5-VL-32B-Instruct} / \texttt{content} / \texttt{text-first} \#1}
\textbf{VL model:} \texttt{Qwen/Qwen2.5-VL-32B-Instruct} \hfill \textbf{Dataset:} \texttt{mscoco} \\[2pt]\textbf{GPT judge:} \texttt{openai/gpt-5.4-mini} \hfill \textbf{Intervention:} \texttt{content}, \(\texttt{layer\_depth}=0.0\) \\[2pt]\textbf{Prompt order} \texttt{text-first}, \textbf{dataset\_index} \texttt{78}

\medskip
\begin{tabularx}{\linewidth}{@{}lX@{}}
\textbf{Gold image caption (ex.)} & \emph{A set of three pizzas in a display case.next to desserts.} \\[2pt]
\textbf{Paired caption} & \emph{The people are walking down the aisles at the super market.} \\[2pt]
\textbf{Image-target · baseline} & \texttt{image} (\textit{outcome}: \texttt{correct}). There are pizzas on display in a supermarket. \\[2pt]
\textbf{Image-target · intervention} & \texttt{neither} (\textit{outcome}: \texttt{neither}). of a with a and on top. \\[2pt]
\textbf{Text-target · baseline} & \texttt{text} (\textit{outcome}: \texttt{correct}). People are walking down the aisles at the supermarket. \\[2pt]
\textbf{Text-target · intervention} & \texttt{image} (\textit{outcome}: \texttt{misled}). of pizzas on trays a display. \\[2pt]
\end{tabularx}
\end{modbox}

\begin{modbox}{\texttt{google/gemma-3-12b-it} / \texttt{marker} / \texttt{image-first} \#1}
\textbf{VL model:} \texttt{google/gemma-3-12b-it} \hfill \textbf{Dataset:} \texttt{mscoco} \\[2pt]\textbf{GPT judge:} \texttt{openai/gpt-5.4-mini} \hfill \textbf{Intervention:} \texttt{marker}, \(\texttt{layer\_depth}=0.0\) \\[2pt]\textbf{Prompt order} \texttt{image-first}, \textbf{dataset\_index} \texttt{93}

\medskip
\begin{tabularx}{\linewidth}{@{}lX@{}}
\textbf{Gold image caption (ex.)} & \emph{A man sitting on a train next to a woman.} \\[2pt]
\textbf{Paired caption} & \emph{A yellow bus waiting at bus stop near some benches.} \\[2pt]
\textbf{Image-target · baseline} & \texttt{image} (\textit{outcome}: \texttt{correct}). The image shows a couple enjoying a bento box lunch on a train. \\[2pt]
\textbf{Image-target · intervention} & \texttt{text} (\textit{outcome}: \texttt{misled}). Yellow waiting at stop benches some. \\[2pt]
\textbf{Text-target · baseline} & \texttt{text} (\textit{outcome}: \texttt{correct}). The caption says "A yellow bus waiting at bus stop near some benches." \\[2pt]
\textbf{Text-target · intervention} & \texttt{image} (\textit{outcome}: \texttt{misled}). A couple eating on train. A couple eating on train. \\[2pt]
\end{tabularx}
\end{modbox}

\begin{modbox}{\texttt{google/gemma-3-12b-it} / \texttt{marker} / \texttt{text-first} \#1}
\textbf{VL model:} \texttt{google/gemma-3-12b-it} \hfill \textbf{Dataset:} \texttt{mscoco} \\[2pt]\textbf{GPT judge:} \texttt{openai/gpt-5.4-mini} \hfill \textbf{Intervention:} \texttt{marker}, \(\texttt{layer\_depth}=0.0\) \\[2pt]\textbf{Prompt order} \texttt{text-first}, \textbf{dataset\_index} \texttt{73}

\medskip
\begin{tabularx}{\linewidth}{@{}lX@{}}
\textbf{Gold image caption (ex.)} & \emph{A group of two women and an older woman between them.} \\[2pt]
\textbf{Paired caption} & \emph{Close up of a white kitchen setup with a coffee maker on counter.} \\[2pt]
\textbf{Image-target · baseline} & \texttt{image} (\textit{outcome}: \texttt{correct}). The image shows three women standing in front of a colorful wall with a banner that reads "Chiquita Anana." \\[2pt]
\textbf{Image-target · intervention} & \texttt{text} (\textit{outcome}: \texttt{misled}). Up of white kitchen a maker a counter.

Three standing a a with a of. A banner that says “” in.

A of with a on.

A of with a on.

A of with a on. \\[2pt]
\textbf{Text-target · baseline} & \texttt{text} (\textit{outcome}: \texttt{correct}). The caption describes a close-up of a white kitchen setup with a coffee maker on the counter. \\[2pt]
\textbf{Text-target · intervention} & \texttt{image} (\textit{outcome}: \texttt{misled}). A of women standing a a wall with a banner.

A of women standing a wall a banner.
A of women standing a a with a banner.
A of women standing a a with banner.
A of women standing a a with banner.
A of women standing a a with banner. \\[2pt]
\end{tabularx}
\end{modbox}

\begin{modbox}{\texttt{google/gemma-3-12b-it} / \texttt{content} / \texttt{image-first} \#1}
\textbf{VL model:} \texttt{google/gemma-3-12b-it} \hfill \textbf{Dataset:} \texttt{mscoco} \\[2pt]\textbf{GPT judge:} \texttt{openai/gpt-5.4-mini} \hfill \textbf{Intervention:} \texttt{content}, \(\texttt{layer\_depth}=0.0\) \\[2pt]\textbf{Prompt order} \texttt{image-first}, \textbf{dataset\_index} \texttt{171}

\medskip
\begin{tabularx}{\linewidth}{@{}lX@{}}
\textbf{Gold image caption (ex.)} & \emph{The orange cat is sleepy on the window sill.} \\[2pt]
\textbf{Paired caption} & \emph{A group of people gathered at a table in a restaurant.} \\[2pt]
\textbf{Image-target · baseline} & \texttt{image} (\textit{outcome}: \texttt{correct}). The image shows a relaxed, light-orange cat lounging on a sunny windowsill. \\[2pt]
\textbf{Image-target · intervention} & \texttt{text} (\textit{outcome}: \texttt{misled}). A of gathered a table a in restaurant. \\[2pt]
\textbf{Text-target · baseline} & \texttt{text} (\textit{outcome}: \texttt{correct}). The caption describes a group of people at a restaurant table. \\[2pt]
\textbf{Text-target · intervention} & \texttt{image} (\textit{outcome}: \texttt{misled}). The is a lying on a. \\[2pt]
\end{tabularx}
\end{modbox}

\begin{modbox}{\texttt{google/gemma-3-12b-it} / \texttt{content} / \texttt{text-first} \#1}
\textbf{VL model:} \texttt{google/gemma-3-12b-it} \hfill \textbf{Dataset:} \texttt{mscoco} \\[2pt]\textbf{GPT judge:} \texttt{openai/gpt-5.4-mini} \hfill \textbf{Intervention:} \texttt{content}, \(\texttt{layer\_depth}=0.0\) \\[2pt]\textbf{Prompt order} \texttt{text-first}, \textbf{dataset\_index} \texttt{97}

\medskip
\begin{tabularx}{\linewidth}{@{}lX@{}}
\textbf{Gold image caption (ex.)} & \emph{Several teddy bears appear to have a picnic on the grass.} \\[2pt]
\textbf{Paired caption} & \emph{A man flying through he air while riding.} \\[2pt]
\textbf{Image-target · baseline} & \texttt{image} (\textit{outcome}: \texttt{correct}). The image shows three teddy bears having a picnic on a red and green plaid blanket with plates and a basket nearby. \\[2pt]
\textbf{Image-target · intervention} & \texttt{text} (\textit{outcome}: \texttt{misled}). A flying through he while riding. \\[2pt]
\textbf{Text-target · baseline} & \texttt{text} (\textit{outcome}: \texttt{correct}). The caption describes a man flying through the air while riding. \\[2pt]
\textbf{Text-target · intervention} & \texttt{image} (\textit{outcome}: \texttt{misled}). The is a picnic with bears. \\[2pt]
\end{tabularx}
\end{modbox}

\begin{modbox}{\texttt{OpenGVLab/InternVL3-14B-hf} / \texttt{marker} / \texttt{image-first} \#1}
\textbf{VL model:} \texttt{OpenGVLab/InternVL3-14B-hf} \hfill \textbf{Dataset:} \texttt{mscoco} \\[2pt]\textbf{GPT judge:} \texttt{openai/gpt-5.4-mini} \hfill \textbf{Intervention:} \texttt{marker}, \(\texttt{layer\_depth}=0.0\) \\[2pt]\textbf{Prompt order} \texttt{image-first}, \textbf{dataset\_index} \texttt{17}

\medskip
\begin{tabularx}{\linewidth}{@{}lX@{}}
\textbf{Gold image caption (ex.)} & \emph{A woman flies a kite near a young child.} \\[2pt]
\textbf{Paired caption} & \emph{A hot dog is smothered with bacon, onions, and mustard.} \\[2pt]
\textbf{Image-target · baseline} & \texttt{image} (\textit{outcome}: \texttt{correct}). The image shows a woman and a child flying a kite on a grassy hill with a tall stone tower in the background. The sky is partly cloudy. \\[2pt]
\textbf{Image-target · intervention} & \texttt{text} (\textit{outcome}: \texttt{misled}). A dog smothered bacon onions and. \\[2pt]
\textbf{Text-target · baseline} & \texttt{text} (\textit{outcome}: \texttt{correct}). A hot dog with bacon, onions, and mustard. \\[2pt]
\textbf{Text-target · intervention} & \texttt{image} (\textit{outcome}: \texttt{misled}). A and a flying a kite a. \\[2pt]
\end{tabularx}
\end{modbox}

\begin{modbox}{\texttt{OpenGVLab/InternVL3-14B-hf} / \texttt{marker} / \texttt{text-first} \#1}
\textbf{VL model:} \texttt{OpenGVLab/InternVL3-14B-hf} \hfill \textbf{Dataset:} \texttt{mscoco} \\[2pt]\textbf{GPT judge:} \texttt{openai/gpt-5.4-mini} \hfill \textbf{Intervention:} \texttt{marker}, \(\texttt{layer\_depth}=0.0\) \\[2pt]\textbf{Prompt order} \texttt{text-first}, \textbf{dataset\_index} \texttt{78}

\medskip
\begin{tabularx}{\linewidth}{@{}lX@{}}
\textbf{Gold image caption (ex.)} & \emph{A set of three pizzas in a display case.next to desserts.} \\[2pt]
\textbf{Paired caption} & \emph{The people are walking down the aisles at the super market.} \\[2pt]
\textbf{Image-target · baseline} & \texttt{image} (\textit{outcome}: \texttt{correct}). The image shows a display case in a pizzeria with various types of pizzas and calzones. There are whole pizzas with different toppings, slices of pizza, and wrapped calzones. The display case has labels indicating the types of items, such as "Meatball Parm," "Hero Rolls," and "Steak \\[2pt]
\textbf{Image-target · intervention} & \texttt{text} (\textit{outcome}: \texttt{misled}). The are walking the at super. \\[2pt]
\textbf{Text-target · baseline} & \texttt{text} (\textit{outcome}: \texttt{correct}). The caption describes a scene in a supermarket where people are walking down the aisles. However, the image provided shows a display case with various types of pizzas and other food items, such as meatball parmas and hero rolls, which are wrapped in pizza dough. There is no direct correlation between the caption and the image \\[2pt]
\textbf{Text-target · intervention} & \texttt{image} (\textit{outcome}: \texttt{misled}). A of pizzas and rolls a. \\[2pt]
\end{tabularx}
\end{modbox}

\begin{modbox}{\texttt{OpenGVLab/InternVL3-14B-hf} / \texttt{content} / \texttt{image-first} \#1}
\textbf{VL model:} \texttt{OpenGVLab/InternVL3-14B-hf} \hfill \textbf{Dataset:} \texttt{mscoco} \\[2pt]\textbf{GPT judge:} \texttt{openai/gpt-5.4-mini} \hfill \textbf{Intervention:} \texttt{content}, \(\texttt{layer\_depth}=0.0\) \\[2pt]\textbf{Prompt order} \texttt{image-first}, \textbf{dataset\_index} \texttt{76}

\medskip
\begin{tabularx}{\linewidth}{@{}lX@{}}
\textbf{Gold image caption (ex.)} & \emph{A white plate topped with food next to a plate of onion rings.} \\[2pt]
\textbf{Paired caption} & \emph{Two signs with arrows pointing to each other for detour.} \\[2pt]
\textbf{Image-target · baseline} & \texttt{image} (\textit{outcome}: \texttt{correct}). A plate of food with fish in creamy sauce, vegetables, and onion rings, a cup of coffee, and a slice of cake. \\[2pt]
\textbf{Image-target · intervention} & \texttt{neither} (\textit{outcome}: \texttt{neither}). A of with a of and a of. \\[2pt]
\textbf{Text-target · baseline} & \texttt{text} (\textit{outcome}: \texttt{correct}). The caption says "Two signs with arrows pointing to each other for detour." \\[2pt]
\textbf{Text-target · intervention} & \texttt{image} (\textit{outcome}: \texttt{misled}). A of food on table with of. \\[2pt]
\end{tabularx}
\end{modbox}

\begin{modbox}{\texttt{OpenGVLab/InternVL3-14B-hf} / \texttt{content} / \texttt{text-first} \#1}
\textbf{VL model:} \texttt{OpenGVLab/InternVL3-14B-hf} \hfill \textbf{Dataset:} \texttt{mscoco} \\[2pt]\textbf{GPT judge:} \texttt{openai/gpt-5.4-mini} \hfill \textbf{Intervention:} \texttt{content}, \(\texttt{layer\_depth}=0.0\) \\[2pt]\textbf{Prompt order} \texttt{text-first}, \textbf{dataset\_index} \texttt{202}

\medskip
\begin{tabularx}{\linewidth}{@{}lX@{}}
\textbf{Gold image caption (ex.)} & \emph{A long train traveling through a train yard.} \\[2pt]
\textbf{Paired caption} & \emph{The empty bench is sitting next to the potted tree.} \\[2pt]
\textbf{Image-target · baseline} & \texttt{image} (\textit{outcome}: \texttt{correct}). The image shows a train station with a train on the tracks, a building labeled "TOPEKA," and several parked cars in the lot. There are multiple train tracks and industrial buildings in the background. \\[2pt]
\textbf{Image-target · intervention} & \texttt{neither} (\textit{outcome}: \texttt{neither}). A of a with a in the. \\[2pt]
\textbf{Text-target · baseline} & \texttt{text} (\textit{outcome}: \texttt{correct}). The caption describes an empty bench sitting next to a potted tree. \\[2pt]
\textbf{Text-target · intervention} & \texttt{image} (\textit{outcome}: \texttt{misled}). A train yard with trains parked the. \\[2pt]
\end{tabularx}
\end{modbox}

%% file: latex/model_output_examples.tex

\begin{modbox}{GPT predicted label: \texttt{image} (closer to gold image captions)}
\textbf{Model:} \texttt{Salesforce/instructblip-vicuna-7b} \hfill \textbf{Dataset:} \texttt{flickr30k} \\[2pt]\textbf{GPT judge:} \texttt{openai/gpt-5.4-mini}

\medskip
\begin{tabularx}{\linewidth}{@{}lX@{}}
\textbf{Target modality} & \texttt{image} \\[2pt]
\textbf{GPT label} & \texttt{image} \quad (\textit{outcome}: \texttt{correct}) \\[2pt]
\textbf{Gold image caption} & \emph{A very aged woman wearing blue is standing in front of a pile of things many in black trash bags.} \\[2pt]
\textbf{Paired caption} & \emph{a girl in a green uniform and a girl in a white uniform are holding lacrosse sticks.} \\[2pt]
\textbf{Model answer} & a girl in a blue uniform is holding a banana
\end{tabularx}
\end{modbox}
\begin{modbox}{GPT predicted label: \texttt{text} (closer to paired caption)}
\textbf{Model:} \texttt{Salesforce/instructblip-vicuna-7b} \hfill \textbf{Dataset:} \texttt{flickr30k} \\[2pt]\textbf{GPT judge:} \texttt{openai/gpt-5.4-mini}

\medskip
\begin{tabularx}{\linewidth}{@{}lX@{}}
\textbf{Target modality} & \texttt{image} \\[2pt]
\textbf{GPT label} & \texttt{text} \quad (\textit{outcome}: \texttt{misled}) \\[2pt]
\textbf{Gold image caption} & \emph{A man in a leather coat plays a saxophone into a microphone.} \\[2pt]
\textbf{Paired caption} & \emph{the man is wearing a backpack and sunglasses while walking through an outdoor market.} \\[2pt]
\textbf{Model answer} & the man is wearing a backpack and sunglasses while walking through an outdoor market
\end{tabularx}
\end{modbox}
\begin{modbox}{GPT predicted label: \texttt{neither}}
\textbf{Model:} \texttt{Salesforce/instructblip-vicuna-7b} \hfill \textbf{Dataset:} \texttt{flickr30k} \\[2pt]\textbf{GPT judge:} \texttt{openai/gpt-5.4-mini}

\medskip
\begin{tabularx}{\linewidth}{@{}lX@{}}
\textbf{Target modality} & \texttt{image} \\[2pt]
\textbf{GPT label} & \texttt{neither} \quad (\textit{outcome}: \texttt{neither}) \\[2pt]
\textbf{Gold image caption} & \emph{Two young women are embracing at the side of a road and looking at an mp3 player.} \\[2pt]
\textbf{Paired caption} & \emph{A small car decorated with flames and other paint is driving amongst a crowd with a tub and flags on the roof.} \\[2pt]
\textbf{Model answer} & carnival
\end{tabularx}
\end{modbox}


\begin{modbox}{GPT predicted label: \texttt{image} (closer to gold image captions)}
\textbf{Model:} \texttt{Salesforce/instructblip-vicuna-7b} \hfill \textbf{Dataset:} \texttt{flickr30k} \\[2pt]\textbf{GPT judge:} \texttt{openai/gpt-5.4-mini}

\medskip
\begin{tabularx}{\linewidth}{@{}lX@{}}
\textbf{Target modality} & \texttt{text} \\[2pt]
\textbf{GPT label} & \texttt{image} \quad (\textit{outcome}: \texttt{misled}) \\[2pt]
\textbf{Gold image caption} & \emph{Girl in black pants and an orange top jumping with one leg outstreached and foot pointed other in background dressed the same.} \\[2pt]
\textbf{Paired caption} & \emph{Three people are having a conversation in a living room.} \\[2pt]
\textbf{Model answer} & people
\end{tabularx}
\end{modbox}

\begin{modbox}{GPT predicted label: \texttt{text} (closer to paired caption)}
\textbf{Model:} \texttt{Salesforce/instructblip-vicuna-7b} \hfill \textbf{Dataset:} \texttt{flickr30k} \\[2pt]\textbf{GPT judge:} \texttt{openai/gpt-5.4-mini}

\medskip
\begin{tabularx}{\linewidth}{@{}lX@{}}
\textbf{Target modality} & \texttt{text} \\[2pt]
\textbf{GPT label} & \texttt{text} \quad (\textit{outcome}: \texttt{correct}) \\[2pt]
\textbf{Gold image caption} & \emph{Brown horse standing in the background while man in a white cowboy hat and blue long-sleeve shirt is walking off.} \\[2pt]
\textbf{Paired caption} & \emph{A brown and white dog jumps in the air to catch a snowball.} \\[2pt]
\textbf{Model answer} & dog
\end{tabularx}
\end{modbox}
\begin{modbox}{GPT predicted label: \texttt{neither}}
\textbf{Model:} \texttt{Salesforce/instructblip-vicuna-7b} \hfill \textbf{Dataset:} \texttt{flickr30k} \\[2pt]\textbf{GPT judge:} \texttt{openai/gpt-5.4-mini}

\medskip
\begin{tabularx}{\linewidth}{@{}lX@{}}
\textbf{Target modality} & \texttt{text} \\[2pt]
\textbf{GPT label} & \texttt{neither} \quad (\textit{outcome}: \texttt{neither}) \\[2pt]
\textbf{Gold image caption} & \emph{A man wearing a blue t-shirt and shorts is holding a newborn baby who is yawning.} \\[2pt]
\textbf{Paired caption} & \emph{A dog playing with a purple toy.} \\[2pt]
\textbf{Model answer} & nothing
\end{tabularx}
\end{modbox}

\begin{modbox}{GPT predicted label: \texttt{image} (closer to gold image captions)}
\textbf{Model:} \texttt{llava-hf/llava-1.5-7b-hf} \hfill \textbf{Dataset:} \texttt{flickr30k} \\[2pt]\textbf{GPT judge:} \texttt{openai/gpt-5.4-mini}

\medskip
\begin{tabularx}{\linewidth}{@{}lX@{}}
\textbf{Target modality} & \texttt{image} \\[2pt]
\textbf{GPT label} & \texttt{image} \quad (\textit{outcome}: \texttt{correct}) \\[2pt]
\textbf{Gold image caption} & \emph{A young man wearing a bandanna on his head stands leaning on a short brick wall with flowers behind him.} \\[2pt]
\textbf{Paired caption} & \emph{A group of people in mostly casual dress meet at an L-shaped table with laptops and notebooks.} \\[2pt]
\textbf{Model answer} & a man wearing a white tank top is sitting on a brick wall
\end{tabularx}
\end{modbox}
\begin{modbox}{GPT predicted label: \texttt{text} (closer to paired caption)}
\textbf{Model:} \texttt{llava-hf/llava-1.5-7b-hf} \hfill \textbf{Dataset:} \texttt{flickr30k} \\[2pt]\textbf{GPT judge:} \texttt{openai/gpt-5.4-mini}

\medskip
\begin{tabularx}{\linewidth}{@{}lX@{}}
\textbf{Target modality} & \texttt{image} \\[2pt]
\textbf{GPT label} & \texttt{text} \quad (\textit{outcome}: \texttt{misled}) \\[2pt]
\textbf{Gold image caption} & \emph{The lady in the black dress is standing near the edge of the cliff taking pictures.} \\[2pt]
\textbf{Paired caption} & \emph{Two girls with brown hair are taking a picture of themselves.} \\[2pt]
\textbf{Model answer} & two girls are taking a picture of themselves
\end{tabularx}
\end{modbox}
\begin{modbox}{GPT predicted label: \texttt{neither}}
\textbf{Model:} \texttt{llava-hf/llava-1.5-7b-hf} \hfill \textbf{Dataset:} \texttt{flickr30k} \\[2pt]\textbf{GPT judge:} \texttt{openai/gpt-5.4-mini}

\medskip
\begin{tabularx}{\linewidth}{@{}lX@{}}
\textbf{Target modality} & \texttt{image} \\[2pt]
\textbf{GPT label} & \texttt{neither} \quad (\textit{outcome}: \texttt{neither}) \\[2pt]
\textbf{Gold image caption} & \emph{Two young girls are arm wrestling in their hotel room while another girl watches.} \\[2pt]
\textbf{Paired caption} & \emph{A young female wearing sandals is in a courtyard and is wearing a colorful backpack and green pants.} \\[2pt]
\textbf{Model answer} & a girl
\end{tabularx}
\end{modbox}

\begin{modbox}{GPT predicted label: \texttt{image} (closer to gold image captions)}
\textbf{Model:} \texttt{llava-hf/llava-1.5-7b-hf} \hfill \textbf{Dataset:} \texttt{flickr30k} \\[2pt]\textbf{GPT judge:} \texttt{openai/gpt-5.4-mini}

\medskip
\begin{tabularx}{\linewidth}{@{}lX@{}}
\textbf{Target modality} & \texttt{text} \\[2pt]
\textbf{GPT label} & \texttt{image} \quad (\textit{outcome}: \texttt{misled}) \\[2pt]
\textbf{Gold image caption} & \emph{Man with blue plaid shirt and white pants skateboards through a building that has graffiti art.} \\[2pt]
\textbf{Paired caption} & \emph{A group of elderly people stand around a booth at a food co-op.} \\[2pt]
\textbf{Model answer} & a man riding a skateboard
\end{tabularx}
\end{modbox}
\begin{modbox}{GPT predicted label: \texttt{text} (closer to paired caption)}
\textbf{Model:} \texttt{llava-hf/llava-1.5-7b-hf} \hfill \textbf{Dataset:} \texttt{flickr30k} \\[2pt]\textbf{GPT judge:} \texttt{openai/gpt-5.4-mini}

\medskip
\begin{tabularx}{\linewidth}{@{}lX@{}}
\textbf{Target modality} & \texttt{text} \\[2pt]
\textbf{GPT label} & \texttt{text} \quad (\textit{outcome}: \texttt{correct}) \\[2pt]
\textbf{Gold image caption} & \emph{A group of people in blue Arabian robes and white headdresses is on yellow bicycles in the street in front of a tall building.} \\[2pt]
\textbf{Paired caption} & \emph{Three men in work-related uniforms standing next to a building.} \\[2pt]
\textbf{Model answer} & three men in work-related uniforms standing next to a building
\end{tabularx}
\end{modbox}
\begin{modbox}{GPT predicted label: \texttt{neither}}
\textbf{Model:} \texttt{llava-hf/llava-1.5-7b-hf} \hfill \textbf{Dataset:} \texttt{flickr30k} \\[2pt]\textbf{GPT judge:} \texttt{openai/gpt-5.4-mini}

\medskip
\begin{tabularx}{\linewidth}{@{}lX@{}}
\textbf{Target modality} & \texttt{text} \\[2pt]
\textbf{GPT label} & \texttt{neither} \quad (\textit{outcome}: \texttt{neither}) \\[2pt]
\textbf{Gold image caption} & \emph{A man in a dark suit talks with a middle-aged man and woman on a stage in front of a large logo of the letter " D " with the numeral five within it.} \\[2pt]
\textbf{Paired caption} & \emph{One man is helping another man with red shorts do a back flip in the yard.} \\[2pt]
\textbf{Model answer} & nothing
\end{tabularx}
\end{modbox}

\begin{modbox}{GPT predicted label: \texttt{image} (closer to gold image captions)}
\textbf{Model:} \texttt{Qwen/Qwen2.5-VL-32B-Instruct} \hfill \textbf{Dataset:} \texttt{flickr30k} \\[2pt]\textbf{GPT judge:} \texttt{openai/gpt-5.4-mini}

\medskip
\begin{tabularx}{\linewidth}{@{}lX@{}}
\textbf{Target modality} & \texttt{image} \\[2pt]
\textbf{GPT label} & \texttt{image} \quad (\textit{outcome}: \texttt{correct}) \\[2pt]
\textbf{Gold image caption} & \emph{Racers run for the finish line in an indoor meet and one man has a big lead.} \\[2pt]
\textbf{Paired caption} & \emph{A young woman in a pink shirt and jeans is walking past some trees.} \\[2pt]
\textbf{Model answer} & the image shows athletes competing in an indoor track event
\end{tabularx}
\end{modbox}
\begin{modbox}{GPT predicted label: \texttt{text} (closer to paired caption)}
\textbf{Model:} \texttt{Qwen/Qwen2.5-VL-32B-Instruct} \hfill \textbf{Dataset:} \texttt{flickr30k} \\[2pt]\textbf{GPT judge:} \texttt{openai/gpt-5.4-mini}

\medskip
\begin{tabularx}{\linewidth}{@{}lX@{}}
\textbf{Target modality} & \texttt{image} \\[2pt]
\textbf{GPT label} & \texttt{text} \quad (\textit{outcome}: \texttt{misled}) \\[2pt]
\textbf{Gold image caption} & \emph{Female football fan wearing brown tank top and blue jean shorts attends a football game in a stadium.} \\[2pt]
\textbf{Paired caption} & \emph{Two blond women are smiling while one is holding a camera to capture their image.} \\[2pt]
\textbf{Model answer} & two blond women are smiling while one is holding a camera to capture their image
\end{tabularx}
\end{modbox}
\begin{modbox}{GPT predicted label: \texttt{neither}}
\textbf{Model:} \texttt{Qwen/Qwen2.5-VL-32B-Instruct} \hfill \textbf{Dataset:} \texttt{flickr30k} \\[2pt]\textbf{GPT judge:} \texttt{openai/gpt-5.4-mini}

\medskip
\begin{tabularx}{\linewidth}{@{}lX@{}}
\textbf{Target modality} & \texttt{image} \\[2pt]
\textbf{GPT label} & \texttt{neither} \quad (\textit{outcome}: \texttt{neither}) \\[2pt]
\textbf{Gold image caption} & \emph{A woman in a blue dress and rain boots mops the cobblestone street.} \\[2pt]
\textbf{Paired caption} & \emph{People are walking in front of a tourist information building displaying the following on its facade : pain is temporary quitting lasts forever.} \\[2pt]
\textbf{Model answer} & a person wearing blue is sweeping the street while people walk by a tourist information building
\end{tabularx}
\end{modbox}

\begin{modbox}{GPT predicted label: \texttt{image} (closer to gold image captions)}
\textbf{Model:} \texttt{Qwen/Qwen2.5-VL-32B-Instruct} \hfill \textbf{Dataset:} \texttt{flickr30k} \\[2pt]\textbf{GPT judge:} \texttt{openai/gpt-5.4-mini}

\medskip
\begin{tabularx}{\linewidth}{@{}lX@{}}
\textbf{Target modality} & \texttt{text} \\[2pt]
\textbf{GPT label} & \texttt{image} \quad (\textit{outcome}: \texttt{misled}) \\[2pt]
\textbf{Gold image caption} & \emph{People are walking in front of a tourist information building displaying the following on its facade : pain is temporary quitting lasts forever.} \\[2pt]
\textbf{Paired caption} & \emph{A woman with blond-hair and a white shirt and white apron is cooking.} \\[2pt]
\textbf{Model answer} & the caption says, "pain is temporary, quitting lasts forever."
\end{tabularx}
\end{modbox}
\begin{modbox}{GPT predicted label: \texttt{text} (closer to paired caption)}
\textbf{Model:} \texttt{Qwen/Qwen2.5-VL-32B-Instruct} \hfill \textbf{Dataset:} \texttt{flickr30k} \\[2pt]\textbf{GPT judge:} \texttt{openai/gpt-5.4-mini}

\medskip
\begin{tabularx}{\linewidth}{@{}lX@{}}
\textbf{Target modality} & \texttt{text} \\[2pt]
\textbf{GPT label} & \texttt{text} \quad (\textit{outcome}: \texttt{correct}) \\[2pt]
\textbf{Gold image caption} & \emph{Two gentle of asian descent sitting at a table or desk seemingly to be concentrating on a project the two are doing together.} \\[2pt]
\textbf{Paired caption} & \emph{There is a gray-haired man wearing a black and orange jacket sitting on chair on the sidewalk playing an accordion.} \\[2pt]
\textbf{Model answer} & a gray-haired man wearing a black and orange jacket sits on a chair on the sidewalk playing an accordion
\end{tabularx}
\end{modbox}
\begin{modbox}{GPT predicted label: \texttt{neither}}
\textbf{Model:} \texttt{Qwen/Qwen2.5-VL-32B-Instruct} \hfill \textbf{Dataset:} \texttt{flickr30k} \\[2pt]\textbf{GPT judge:} \texttt{openai/gpt-5.4-mini}

\medskip
\begin{tabularx}{\linewidth}{@{}lX@{}}
\textbf{Target modality} & \texttt{text} \\[2pt]
\textbf{GPT label} & \texttt{neither} \quad (\textit{outcome}: \texttt{neither}) \\[2pt]
\textbf{Gold image caption} & \emph{Two women with white and purple gowns while posing for the camera.} \\[2pt]
\textbf{Paired caption} & \emph{An Asian man sells a hot confection from his stand in a crowded market.} \\[2pt]
\textbf{Model answer} & a caption is a brief description of an image or scene
\end{tabularx}
\end{modbox}

\begin{modbox}{GPT predicted label: \texttt{image} (closer to gold image captions)}
\textbf{Model:} \texttt{google/gemma-3-12b-it} \hfill \textbf{Dataset:} \texttt{flickr30k} \\[2pt]\textbf{GPT judge:} \texttt{openai/gpt-5.4-mini}

\medskip
\begin{tabularx}{\linewidth}{@{}lX@{}}
\textbf{Target modality} & \texttt{image} \\[2pt]
\textbf{GPT label} & \texttt{image} \quad (\textit{outcome}: \texttt{correct}) \\[2pt]
\textbf{Gold image caption} & \emph{A man wearing a light blue shirt and gray pants is crouched in a field holding nuts in his hand.} \\[2pt]
\textbf{Paired caption} & \emph{Skateboarder performs a stunt on an uphill residential street.} \\[2pt]
\textbf{Model answer} & the image shows a man crouched in a field, tending to plants and holding a root vegetable in his hand
\end{tabularx}
\end{modbox}
\begin{modbox}{GPT predicted label: \texttt{text} (closer to paired caption)}
\textbf{Model:} \texttt{google/gemma-3-12b-it} \hfill \textbf{Dataset:} \texttt{flickr30k} \\[2pt]\textbf{GPT judge:} \texttt{openai/gpt-5.4-mini}

\medskip
\begin{tabularx}{\linewidth}{@{}lX@{}}
\textbf{Target modality} & \texttt{image} \\[2pt]
\textbf{GPT label} & \texttt{text} \quad (\textit{outcome}: \texttt{misled}) \\[2pt]
\textbf{Gold image caption} & \emph{A woman wearing a blue shirt and high heels stands on the sidewalk next to a man.} \\[2pt]
\textbf{Paired caption} & \emph{A man in jeans and a leather jacket walking by a poster advertising Miller Lite.} \\[2pt]
\textbf{Model answer} & the image shows a woman in a blue dress and a man with headphones standing on a sidewalk next to a store window displaying lingerie
\end{tabularx}
\end{modbox}

\begin{modbox}{GPT predicted label: \texttt{image} (closer to gold image captions)}
\textbf{Model:} \texttt{google/gemma-3-12b-it} \hfill \textbf{Dataset:} \texttt{flickr30k} \\[2pt]\textbf{GPT judge:} \texttt{openai/gpt-5.4-mini}

\medskip
\begin{tabularx}{\linewidth}{@{}lX@{}}
\textbf{Target modality} & \texttt{text} \\[2pt]
\textbf{GPT label} & \texttt{image} \quad (\textit{outcome}: \texttt{misled}) \\[2pt]
\textbf{Gold image caption} & \emph{A man uses two sticks to control two tan goats in a pen while a black and white dog watches.} \\[2pt]
\textbf{Paired caption} & \emph{Two females sitting on large backed stools in front of a counter inside of a house with one female turning her head back.} \\[2pt]
\textbf{Model answer} & the caption describes a scene of a sheep herding demonstration with a handler, a dog, and sheep
\end{tabularx}
\end{modbox}
\begin{modbox}{GPT predicted label: \texttt{text} (closer to paired caption)}
\textbf{Model:} \texttt{google/gemma-3-12b-it} \hfill \textbf{Dataset:} \texttt{flickr30k} \\[2pt]\textbf{GPT judge:} \texttt{openai/gpt-5.4-mini}

\medskip
\begin{tabularx}{\linewidth}{@{}lX@{}}
\textbf{Target modality} & \texttt{text} \\[2pt]
\textbf{GPT label} & \texttt{text} \quad (\textit{outcome}: \texttt{correct}) \\[2pt]
\textbf{Gold image caption} & \emph{A hockey player in black and white collides with another player wearing blue.} \\[2pt]
\textbf{Paired caption} & \emph{A man in a gray shirt sits with several pairs of shoes around him.} \\[2pt]
\textbf{Model answer} & the caption says "a man in a gray shirt sits with several pairs of shoes around him."
\end{tabularx}
\end{modbox}
\begin{modbox}{GPT predicted label: \texttt{neither}}
\textbf{Model:} \texttt{google/gemma-3-12b-it} \hfill \textbf{Dataset:} \texttt{flickr30k} \\[2pt]\textbf{GPT judge:} \texttt{openai/gpt-5.4-mini}

\medskip
\begin{tabularx}{\linewidth}{@{}lX@{}}
\textbf{Target modality} & \texttt{text} \\[2pt]
\textbf{GPT label} & \texttt{neither} \quad (\textit{outcome}: \texttt{neither}) \\[2pt]
\textbf{Gold image caption} & \emph{People viewing a sculpture with a man without a face but with a suit tie glasses and a hat holding a multi color umbrella.} \\[2pt]
\textbf{Paired caption} & \emph{Three men are working on some kind of equipment but I 'm not sure whether it 's a ship or an oil rig.} \\[2pt]
\textbf{Model answer} & the caption says "guts and capture."
\end{tabularx}
\end{modbox}